\documentclass[11pt,letterpaper]{article}

\usepackage[textwidth=6in,textheight=8.5in,centering]{geometry}
\usepackage{amsmath,amssymb,amsthm}
\usepackage{natbib,graphicx,url}
\usepackage[algo2e,ruled]{algorithm2e}
\usepackage[hidelinks]{hyperref}
\bibpunct{(}{)}{;}{a}{,}{,}
\newtheorem{theorem}{Theorem}
\newenvironment{keywords}{\par\small\noindent\textbf{Keywords:}\ }{\par}
\title{Robust Recourse under Causal Graph Ambiguity}
\usepackage{times}
\usepackage{multicol}
\usepackage{multirow}
\usepackage{booktabs}
\usepackage{array}
\usepackage{xcolor}
\hypersetup{hypertexnames=false}
\usepackage{xurl}
\usepackage[normalem]{ulem}

\usepackage{pifont}
\usepackage{placeins}
\newsavebox{\revisiontable}

\usepackage{float}

\usepackage{etoolbox}
\BeforeBeginEnvironment{table}{%
  \ifnum\value{table}=0\relax\FloatBarrier\fi
}

\usepackage{amsmath,amssymb,url}
\usepackage{tikz}
\usetikzlibrary{calc,arrows.meta,positioning,fit,shadows,backgrounds}
\usepackage[capitalise,noabbrev]{cleveref}
\crefname{algocf}{Algorithm}{Algorithms}
\Crefname{algocf}{Algorithm}{Algorithms}
\newtheorem{observation}[theorem]{Observation}
\crefname{observation}{Observation}{Observations}
\Crefname{observation}{Observation}{Observations}

\newcommand{\rev}[1]{#1}

\newcommand{\xF}{\mathbf{x}^{\texttt{F}}}
\newcommand{\xCF}{\mathbf{x}^{\texttt{CF}}}
\newcommand{\XCF}{\mathbf{X}^{\texttt{CF}}}
\newcommand{\thresh}{\texttt{thresh}}
\newcommand{\XI}{\mathbf{X}_{\mathcal{I}}}
\newcommand{\btheta}{\boldsymbol{\theta}}
\newcommand{\AF}{\mathbb{A}^{\texttt{F}}}
\newcommand{\costF}{\mathrm{cost}^{\texttt{F}}}
\newcommand{\GG}{\mathbb{G}}
\newcommand{\ms}[2]{%
  \ensuremath{#1\,{\color{stdgray}\pm #2}}%
}
\definecolor{stdgray}{gray}{0.60}
\definecolor{intv}{HTML}{0072B2}
\definecolor{chg}{HTML}{E69F00}
\definecolor{good}{HTML}{009E73}
\definecolor{bad}{HTML}{D55E00}
\definecolor{amb}{HTML}{CC79A7}

\author{%
  \renewcommand{\arraystretch}{1.05}%
  \begin{tabular}{@{}c@{\hspace{2em}}c@{}}
    Muyan Anna Li\textsuperscript{1,2} & Juliette Sinnott\textsuperscript{3}\\
    {\small\href{mailto:a5li@uwaterloo.ca}{\texttt{a5li@uwaterloo.ca}}} &
    {\small\href{mailto:jssinnott@uwaterloo.ca}{\texttt{jssinnott@uwaterloo.ca}}}\\[0.55em]
    Mohammad Kohandel\textsuperscript{3} & Amir-Hossein Karimi\textsuperscript{1,2,4}\\
    {\small\href{mailto:kohandel@uwaterloo.ca}{\texttt{kohandel@uwaterloo.ca}}} &
    {\small\href{mailto:amirh.karimi@uwaterloo.ca}{\texttt{amirh.karimi@uwaterloo.ca}}}\\[0.75em]
    \multicolumn{2}{c}{\small\textsuperscript{1}David R. Cheriton School of Computer Science, University of Waterloo, Canada}\\
    \multicolumn{2}{c}{\small\textsuperscript{2}Department of Electrical and Computer Engineering, University of Waterloo, Canada}\\
    \multicolumn{2}{c}{\small\textsuperscript{3}Department of Applied Mathematics, University of Waterloo, Canada}\\
    \multicolumn{2}{c}{\small\textsuperscript{4}Vector Institute for Artificial Intelligence, Toronto, Canada}
  \end{tabular}%
}
\date{}
\makeatletter
\renewcommand{\@maketitle}{%
  \newpage\null\vskip 0.5em
  \begin{center}%
    {\LARGE\@title\par}%
    \vskip 1em
    {\normalsize\lineskip 0.3em
      \begin{tabular}[t]{c}\@author\end{tabular}\par}%
  \end{center}%
  \vskip 0.3em
}
\makeatother
\hypersetup{%
  pdftitle={Robust Recourse under Causal Graph Ambiguity},
  pdfauthor={Muyan Anna Li, Juliette Sinnott, Mohammad Kohandel, Amir-Hossein Karimi}
}

\begin{document}

\maketitle

\begin{abstract}

Causal algorithmic recourse recommends actions to reverse unfavorable decisions by modeling how interventions change an individual's features.
Existing methods typically assume one causal graph that describes the system and account for uncertainty in the causal mechanisms.
However, there usually exist several graphs that fit the same data but can predict different effects for the same action.
{Certain actions may achieve recourse across multiple graphs, but as the set of potential graphs grows, a single best action may become infeasible.}

We propose Worst-Case Recourse and Graph-Weighted Recourse to find a low-cost action that succeeds across all candidate graphs or a specified fraction of graph weight, respectively.
Assuming a set of plausible graphs, for each graph we fit Gaussian process structural equations and assess success using the posterior mean--standard-deviation criterion.
We use gradient-based search to determine minimal cost actions and check the validity of the actions on fresh posterior samples.
In synthetic and semi-synthetic benchmarks, relaxing \rev{robustness, the requirement that an action succeed under every candidate graph,} can increase the number of valid recommendations for recourse and lower costs of actions for both methods.
{\leavevmode In the 20-node GP benchmark, relaxation increases the number of factual instances offered recourse \rev{(coverage)} from 341 to 436 out of 1,000}, { while the number \rev{of true successes, actions that flip the decision} under the generating model\rev{,} falls from 251 to 184}, illustrating the \rev{coverage--validity} trade-off.
Accounting for graph ambiguity can help decision-makers recommend actions that remain effective even when the causal mechanisms of the system are not fully known.
\end{abstract}


\section{Introduction}\label{sec:intro}

\suppressfloats[t]% no top floats on page 1, so Figure 1 goes to the top of page 2
\begin{figure}[t]
\centering
\resizebox{0.987525\textwidth}{!}{% Motivating example: mortgage pre-approval under a two-DAG MEC.
% Requires: tikz (arrows.meta, positioning, backgrounds), pifont.
% Colours intv, chg, good, bad, amb are defined in the preamble.
\begin{tikzpicture}[>={Stealth[length=1.6mm]},
 v/.style={draw,circle,minimum size=6.5mm,inner sep=0pt,font=\small},
 im/.style={v,dashed},
 do/.style={v,fill=intv!22,draw=intv,line width=.9pt},
 val/.style={font=\scriptsize,inner sep=1pt},
 cut/.style={draw=black!30,dashed,->},
 amb/.style={draw=amb,line width=1pt},
 ttl/.style={font=\small\bfseries,anchor=west},
 hd/.style={font=\small,align=center},
 rl/.style={font=\small,align=left,anchor=west,text width=38mm},
 ok/.style={font=\scriptsize,align=left,anchor=west},
 every path/.style={line width=.4pt}]

% ---------- The same action under each DAG in the MEC ----------
\node[hd,anchor=west] at (-0.4,-2.6) {\textbf{Action}};
\node[hd] at (5.5,-2.6) {\textbf{Graph A}: $S\to D$ \textcolor{amb}{($-0.8$)}};
\node[hd] at (11.5,-2.6) {\textbf{Graph B}: $D\to S$ \textcolor{amb}{($-0.5$)}};
\draw[black!40] (-0.4,-2.9)--(16.1,-2.9);

\newcommand{\cg}[1]{{\color{chg}\mathbf{#1}}}
% #1 y, #2 tag, #3 x
\newcommand{\minidag}[3]{%
 \node[v] (s#2) at (#3,#1) {$S$};
 \node[v] (d#2) at (#3+1.3,#1) {$D$};
 \node[v] (c#2) at (#3+2.6,#1) {$C$};
 \draw[->] (d#2)--(c#2);}

\foreach \row/\y/\lbl/\cost/\sA/\dA/\cA/\okA/\sB/\dB/\cB/\okB/\doS/\doD in {
  1/-3.55/{Save \$6.25k}/6.25/24.25/\cg{5}/\cg{680}/{\textcolor{good}{\ding{51} approved}}/24.25/10/660/{\textcolor{bad}{\ding{55} denied ($C{<}680$)}}/1/0,
  2/-4.75/{Pay down \$5k}/5/18/5/\cg{680}/{\textcolor{bad}{\ding{55} denied ($S{<}20$)}}/\cg{20.5}/5/\cg{680}/{\textcolor{good}{\ding{51} approved}}/0/1,
  3/-5.95/{Save \$2k \textit{and} pay \$5k}/7/20/5/\cg{680}/{\textcolor{good}{\ding{51} approved}}/20/5/\cg{680}/{\textcolor{good}{\ding{51} approved}}/1/1}{
 \node[rl] at (-0.4,\y) {\lbl\\[-1pt]{\scriptsize cost \cost}};
 \minidag{\y}{A}{4.2}
 \minidag{\y}{B}{10.2}
 % Intervening on the head of the S--D edge cuts it.
 \ifnum\doD=1 \draw[cut] (sA)--(dA); \else \draw[->,amb] (sA)--(dA); \fi
 \ifnum\doS=1 \draw[cut] (dB)--(sB); \else \draw[->,amb] (dB)--(sB); \fi
 \ifnum\doS=1 \node[do] at (sA) {$S$}; \node[do] at (sB) {$S$}; \fi
 \ifnum\doD=1 \node[do] at (dA) {$D$}; \node[do] at (dB) {$D$}; \fi
 \node[val,below=1pt of sA] {$\sA$}; \node[val,below=1pt of dA] {$\dA$}; \node[val,below=1pt of cA] {$\cA$};
 \node[val,below=1pt of sB] {$\sB$}; \node[val,below=1pt of dB] {$\dB$}; \node[val,below=1pt of cB] {$\cB$};
 \node[ok] at (7.15,\y) {\okA};
 \node[ok] at (13.15,\y) {\okB};
 \ifnum\row<3 \draw[black!15] (-0.4,\y-0.6)--(16.1,\y-0.6); \fi
}
\begin{scope}[on background layer]
 \fill[good!10,rounded corners=3pt] (-0.5,-6.6) rectangle (16.2,-5.4);
\end{scope}
\end{tikzpicture}}\par
\caption{Toy example of mortgage recourse under causal ambiguity. Graph A and B are equally likely, however each graph's cheapest action fails under the other. Setting both savings and debt removes this dependence on edge direction so both graphs approve at $L_1$ cost 7. Blue nodes are intervention targets, dashed grey edges are cut by intervention, and orange values mark downstream changes.}
\label{fig:motivation}
\end{figure}
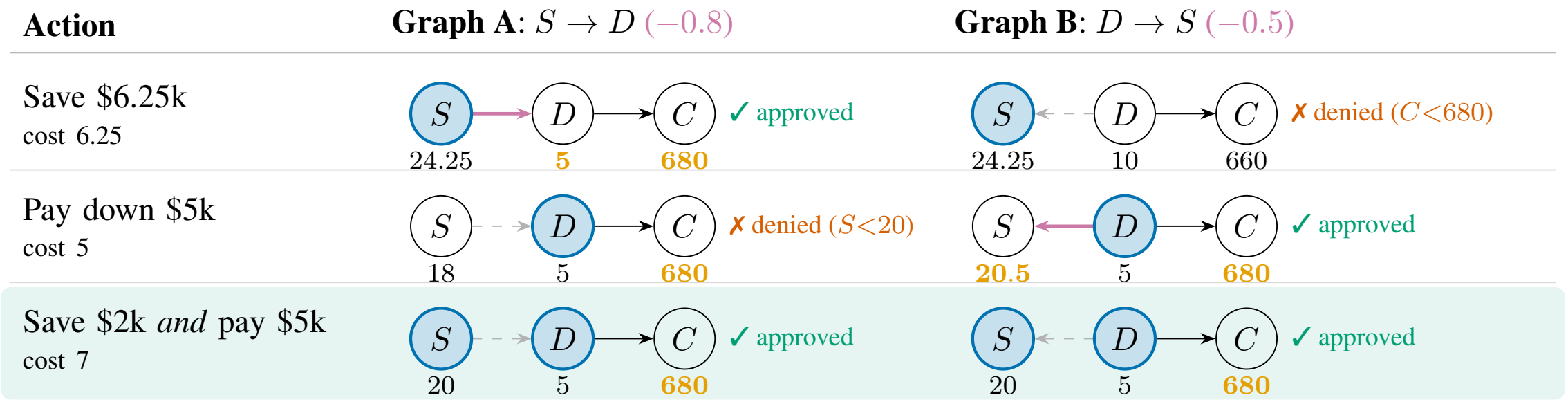

Algorithmic recourse aims to give an individual who receives an unfavorable decision an action that will change it~\citep{wachterCounterfactualExplanationsOpening2017, ustunActionableRecourseLinear2019, venkatasubramanian2020philosophical}; see~\citet{karimi2022survey} and~\citet{verma2024counterfactual} for surveys.  In high-stakes settings such as credit, hiring, and public services, a useful recommendation should be \emph{valid} (it changes the decision), \emph{affordable} (its cost to the individual is low), and \emph{robust} (it remains valid under realistic uncertainty).
\rev{Unlike counterfactual explanations~\citep{martens2014explaining, tolomei2017interpretable, joshi2019realistic, karimi2020model, mothilal2020explaining, poyiadzi2020face}, which change features independently and can ignore causal dependencies between them~\citep{mahajan2019preserving, barocas2020hidden}, causal recourse~\citep{karimi2021algorithmic} models actions as interventions $a=do(\XI=\btheta)$ on a structural causal model (SCM)~\citep{pearl2009causality}, so that an action's downstream effects are accounted for.}

Most causal recourse methods assume that a causal graph is given, even though observational data cannot identify a single valid graph.  Instead, under standard causal identification assumptions, there exists a Markov equivalence class (MEC) of possible graphs~\citep{verma1990equivalence, spirtes2000causation}.  Graphs in an MEC agree on all conditional independences in the data but can differ in edge directions, parent sets, and mediation paths~\citep{andersson1997characterization}.  Two observationally equivalent graphs can therefore imply materially different outcomes for the same intervention~\citep{maathuis2009estimating}: an action optimized for one graph may fail under another in the MEC.

Can we recommend one affordable intervention that changes an unfavorable classifier decision under every graph in an MEC? \Cref{fig:motivation} illustrates this with a mortgage applicant: the data cannot tell whether savings reduce debt or debt prevents saving, so the cheapest action under each graph is denied under the other, while a slightly more expensive action that sets both savings and debt is approved under both. We first ask when \rev{there exists} such an optimal action and what it costs, then we explore the gains and losses from relaxing \rev{the requirement that it succeed under every graph}.

Graph ambiguity is distinct from uncertainty in the structural equations under a fixed graph, which~\citet{karimiAlgorithmicRecourseImperfect2020} address with Gaussian process SCMs (GP-SCMs).  Our method handles both: GP-SCMs capture uncertainty in the structural equations within each graph, while our constraints range over every graph in the MEC.

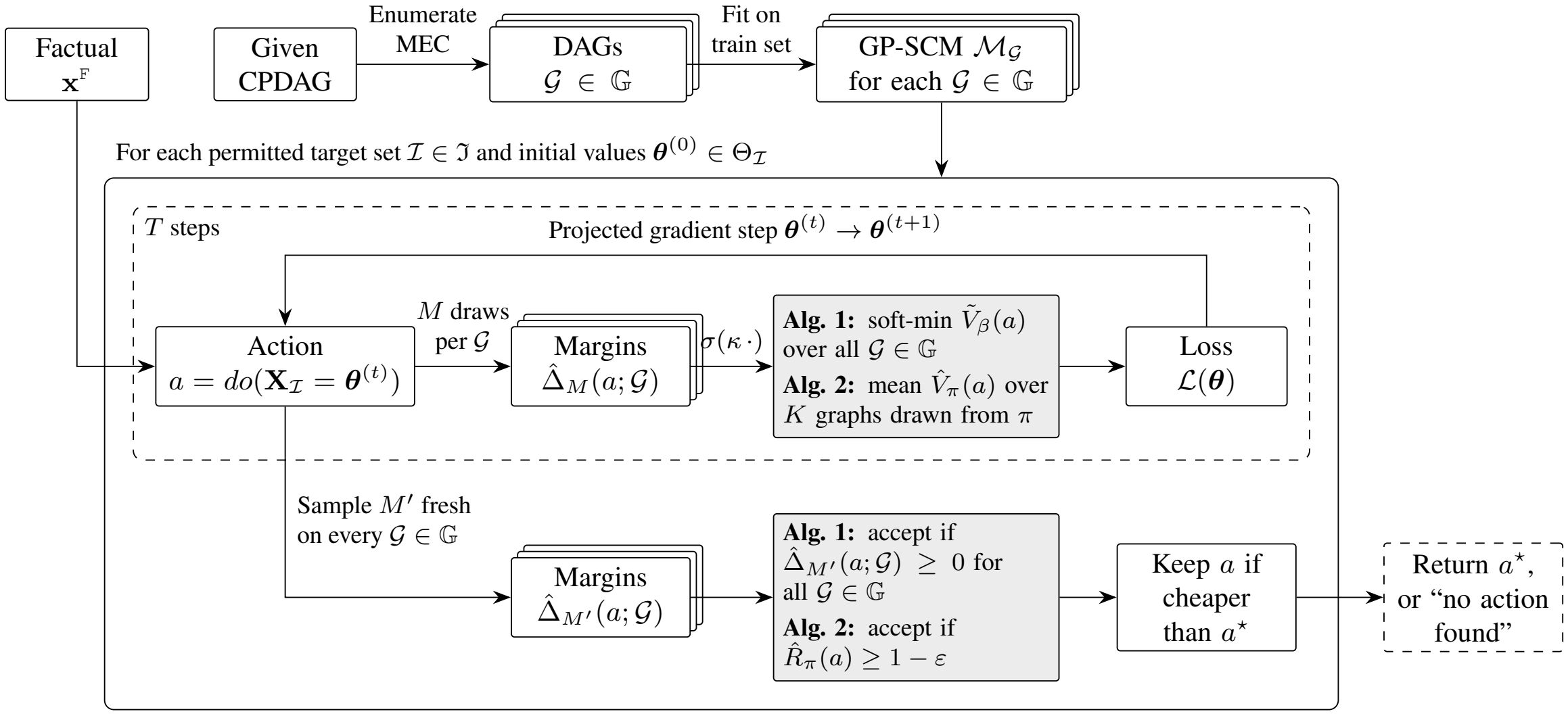
\begin{figure}[t]
\centering
\resizebox{\linewidth}{!}{\begin{tikzpicture}[
  x=0.71cm,
  font=\footnotesize,
  >={Stealth[length=2mm]},
  box/.style={draw, rounded corners=1pt, fill=white, align=center, minimum height=8mm, inner sep=3pt},
  stack/.style={box, double copy shadow={shadow xshift=2pt, shadow yshift=2pt}},
  diff/.style={box, fill=gray!15, align=left, font=\scriptsize, text width=30mm},
  result/.style={box, dashed},
  lab/.style={font=\scriptsize, align=center, midway, above},
  frame/.style={draw, rounded corners=3pt, inner sep=7pt},
]
% --- Offline: MEC enumeration (CPDAG is given) and SCM fitting ---
\node[box, minimum width=16mm] (cpdag) at (0,0) {Given\\CPDAG};
\node[stack, text width=20mm] (dags) at (4.8,0) {DAGs\\$\mathcal{G}\in\GG$};
\node[stack, text width=26mm] (scms) at (10.4,0) {GP-SCM $\mathcal{M}_{\mathcal{G}}$\\for each $\mathcal{G}\in\GG$};
\draw[->] (cpdag) -- node[lab]{Enumerate\\MEC} (dags);
\draw[->] (dags) -- node[lab]{Fit on\\train set} (scms);

% --- Search: gradient steps ---
\node[box, text width=14mm] (fact) at (-3.3,0) {Factual\\$\xF$};
\node[box, text width=27mm] (act) at (0,-3.4) {Action\\$a=do(\XI=\btheta^{(t)})$};
\node[stack, text width=18mm] (marg) at (5.0,-3.4) {Margins\\$\hat{\Delta}_M(a;\mathcal{G})$};
\node[diff] (comb) at (10.0,-3.4) {\textbf{Alg.~1:} soft-min $\tilde{V}_\beta(a)$ over all $\mathcal{G}\in\GG$\\[2pt]\textbf{Alg.~2:} mean $\hat{V}_\pi(a)$ over $K$ graphs drawn from $\pi$};
\node[box, text width=16mm] (loss) at (14.6,-3.4) {Loss\\$\mathcal{L}(\btheta)$};
\draw[->] (fact.south) |- (act.west);
\draw[->] (act) -- node[lab]{$M$ draws\\per $\mathcal{G}$} (marg);
\draw[->] (marg) -- node[lab]{$\sigma(\kappa\,\cdot)$} (comb);
\draw[->] (comb) -- (loss);
\draw[->] (loss.north) -- ++(0,0.8) -| node[lab, pos=0.25]{Projected gradient step $\btheta^{(t)}\to\btheta^{(t+1)}$} (act.north);
\coordinate (gtop) at ([yshift=11mm]loss.north);

% --- Verification of every iterate ---
\node[stack, text width=18mm] (vmarg) at (5.0,-6.0) {Margins\\$\hat{\Delta}_{M'}(a;\mathcal{G})$};
\node[diff] (acc) at (10.0,-6.0) {\textbf{Alg.~1:} accept if $\hat{\Delta}_{M'}(a;\mathcal{G})\geq0$ for all $\mathcal{G}\in\GG$\\[2pt]\textbf{Alg.~2:} accept if $\hat{R}_\pi(a)\geq1-\varepsilon$};
\node[box, text width=18mm] (upd) at (14.6,-6.0) {Keep $a$ if cheaper than $a^\star$};
\draw[->] (act.south) |- node[font=\scriptsize, align=left, right, pos=0.3]{Sample $M'$ fresh\\on every $\mathcal{G}\in\GG$} (vmarg.west);
\draw[->] (vmarg) -- (acc);
\draw[->] (acc) -- (upd);

% --- Frames ---
\begin{scope}[on background layer]
  \node[frame, dashed, fit=(act)(marg)(comb)(loss)(gtop)] (inner) {};
  \node[font=\scriptsize, anchor=north west] at (inner.north west) {$T$ steps};
  \node[frame, fit=(inner)(vmarg)(acc)(upd), inner sep=9pt] (outer) {};
  \node[font=\scriptsize, anchor=south west] at (outer.north west) {For each permitted target set $\mathcal{I}\in\mathfrak{I}$ and initial values $\btheta^{(0)}\in\Theta_{\mathcal{I}}$};
\end{scope}
\draw[->] (scms.south) -- (scms.south |- outer.north);

% --- Output ---
\node[result, text width=18mm, anchor=west] (ret) at ([xshift=5mm]outer.east |- upd) {Return $a^\star$, or ``no action found''};
\draw[->] (upd) -- (ret);
\end{tikzpicture}}
\caption{Overview of \cref{alg:worst-case,alg:weighted}. Top: enumerate the given CPDAG's MEC $\GG$ and fit each graph's GP-SCM on training data. Bottom: $a=do(\XI=\btheta)$ sets targets $\mathcal I\subseteq[d]$ to $\btheta$. For each permitted $\mathcal I\in\mathfrak I$ and initialization $\btheta^{(0)}\in\Theta_{\mathcal I}$, take $T$ gradient steps on $\mathcal L(\btheta)$ using $M$ counterfactual samples per graph; verify each iterate on $M'$ fresh samples per graph. \rev{$\hat{R}_\pi(a)$ is the estimated graph weight on which $a$ is valid (\cref{sec:alg2}).} Stacked boxes are computed per graph; shading marks the two differing steps.}
\label{fig:pipeline}
\end{figure}

Prior work studies robustness to action execution, recorded features, predictive models, and temporal change. See \cref{sec:related}. We study uncertainty about causal graph structure, combining all-model and graph-weighted requirements with GP models for uncertain mechanisms.
\Cref{fig:pipeline} gives an overview: we enumerate MEC DAGs, fit a GP-SCM to each, and seek a low-cost action verified on all graphs (\cref{alg:worst-case}) or a specified fraction of graph weight (\cref{alg:weighted}).

We make the following contributions:
\begin{itemize}
    \item We formulate minimal-cost recourse using one action across every DAG in the MEC \eqref{eq:objective}, and a relaxation requiring approval under a specified fraction of graph weight \eqref{eq:robustness}.
    \item We propose two gradient-based search algorithms (\cref{alg:worst-case,alg:weighted}) with smooth worst-case and graph-weighted objectives, respectively. Both retain the cheapest visited action passing independent Monte Carlo checks.
    \item We fit a GP-SCM to each candidate graph and apply a classifier-score lower confidence bound (LCB) criterion, so returned actions account for uncertainty in both structural equations and graph structure (\cref{sec:background,tab:gp-lcb}).
    \item We prove and illustrate that recourse under some graphs need not admit a common robust action (\cref{prop:impossibility,app:graded-motivation}). Relaxation increases coverage \rev{(the fraction of individuals for whom an action is returned)} and lowers paired costs in our fitted studies, but can sacrifice \rev{true validity} (\cref{tab:ols-results,tab:gp-lcb,app:fitted-paired}).
\end{itemize}
\Cref{sec:background} introduces causal recourse and probabilistic recourse with GP-SCMs, \cref{sec:methodology} formalizes our objective \rev{under a given CPDAG} and presents the two algorithms (\cref{sec:alg1,sec:alg2}), \cref{sec:experiments} reports experiments, and \cref{sec:discussion} discusses limitations.  We review related work on robust recourse and on uncertainty in the causal graph in \cref{sec:related}.

\section{Background}\label{sec:background}

We follow the notation and definitions of recourse in~\citet{karimiAlgorithmicRecourseImperfect2020}, which builds on the causal recourse formulation of~\citet{karimi2021algorithmic}; see also~\citet{vonkugelgen2022fairness}.  Predictions come from a pre-trained model $h$.  For simplicity, we assume that $h$ predicts a single continuous value in $[0,1]$, where scores of at least $0.5$ are desirable and scores below are undesirable.

\paragraph{Actions.}  For any factual instance $\xF$ whose prediction is unfavorable ($h(\xF)<0.5$), recourse seeks an action $a$ such that the classifier predicts a \emph{favorable outcome}: $h(\xCF(a))\geq0.5$.  This decision boundary can be altered depending on the context-specific definition of a favorable outcome.  \emph{Actions} are modeled as interventions in the sense of Pearl's $do$-operator~\citep{pearl1995causal, pearl2009causality}, meaning they describe which features in $\xF$ should be changed and to what values.  In our optimization approaches, we define an action $a$ in terms of its intervention targets $\mathcal{I}\subseteq[d]$ and the intervention values $\btheta$.  With this we can write: $a=do(\XI=\btheta)$.  We denote by $\mathfrak{I}$ the collection of permitted intervention target sets, by $\Theta_{\mathcal{I}}$ the feasible values for target set $\mathcal{I}$, and by $\AF=\{do(\XI=\btheta):\mathcal{I}\in\mathfrak{I},\,\btheta\in\Theta_{\mathcal{I}}\}$ the resulting set of feasible actions for $\xF$.

\paragraph{Structural equations.}  Downstream changes from an action are determined from an SCM over features $\mathbf{X}=\{X_1,\ldots,X_d\}$, whose structural equations $X_r:=f_r(\mathbf{X}_{\mathrm{pa}(r)},U_r)$, $r\in[d]$, describe how each feature is generated from its parents $\mathbf{X}_{\mathrm{pa}(r)}$ in a directed acyclic graph (DAG) $\mathcal{G}$ and an unobserved noise variable $U_r$~\citep{peters2017elements}.  For a fixed graph $\mathcal{G}$,~\citet{karimiAlgorithmicRecourseImperfect2020} fit a GP-SCM $\mathcal{M}_{\mathcal{G}}$ to training data $\{\mathbf{x}^i\}_{i=1}^n$:
\begin{equation}\label{eq:gp-scm}
X_r:=f_r(\mathbf{X}_{\mathrm{pa}_{\mathcal{G}}(r)})+U_r,\qquad f_r\sim\mathcal{GP}(0,k_r),\qquad U_r\sim\mathcal{N}(0,\sigma_r^2),\qquad r\in[d],
\end{equation}
where $\mathrm{pa}_{\mathcal{G}}(r)$ are the parents of $X_r$ in $\mathcal{G}$. The GP priors are independent across nodes and the exogenous noise is assumed to be additive and independent.

\paragraph{Counterfactuals.}  When computing a point-counterfactual $\xCF(a):=\mathbf{x}(a)\mid\xF$ we consider direct changes to features\rev{ (interventions)}, and any downstream changes that arise from them.  \rev{Because GPs quantify uncertainty in the equations, under a GP-SCM we obtain a} \emph{counterfactual random variable} $\XCF(a;\mathcal{G}):=\mathbf{X}(a)\mid\xF,\{\mathbf{x}^i\}_{i=1}^n$ under $\mathcal{M}_{\mathcal{G}}$.  The intervened features are fixed at $\btheta$, and downstream features are sampled in topological order from Gaussian counterfactual conditionals with closed-form means and variances~\citep{karimiAlgorithmicRecourseImperfect2020}.

\paragraph{Validity.}  Because we are using probabilistic counterfactuals, we require the expected classifier score to satisfy $\mathbb{E}\big[h(\XCF(a;\mathcal{G}))\big]\geq\thresh(a;\mathcal{G})$, with
\begin{equation}\label{eq:thresh}
\thresh(a;\mathcal{G})=0.5+\gamma_{\textsc{lcb}}\sqrt{\operatorname{Var}\big[h(\XCF(a;\mathcal{G}))\big]},
\end{equation}
where the expectation and variance are taken over $\XCF(a;\mathcal{G})$. This lower-confidence-bound-style threshold parallels confidence bounds in Gaussian process optimization~\citep{srinivas2010gaussian}: larger $\gamma_{\textsc{lcb}}$ is more conservative, while $\gamma_{\textsc{lcb}}=0$ only requires the expected output to be at least $0.5$.

\paragraph{Cost.}  There are usually infinitely many actions that can achieve recourse\rev{; therefore,} we seek the action of minimal cost $\costF(a)$: defined as the distance between the factual values \rev{$\xF_{\mathcal{I}}$} and the intervened values \rev{$\btheta$}.  Combining this with the criterion above gives the optimization problem:
\begin{equation}\label{eq:prob-recourse}
\min_{a=do(\XI=\btheta)\in\AF} \costF(a) \quad\text{subject to}\quad
 \mathbb{E}\big[h(\XCF(a;\mathcal{G}))\big]\geq\thresh(a;\mathcal{G}).
\end{equation}

\section{Methodology}\label{sec:methodology}
Our formulation requires \emph{validity} (flipping the prediction), \emph{affordability} (minimizing cost), and \emph{robustness} (validity across plausible causal graphs).

\paragraph{Validity: flipping the prediction with an action.}
\rev{For a graph $\mathcal{G}$, we write the probabilistic criterion of \cref{sec:background} as a} per-graph \emph{margin}
\begin{equation}\label{eq:margin}
\Delta(a;\mathcal{G})=\mathbb{E}\big[h(\XCF(a;\mathcal{G}))\big]-\thresh(a;\mathcal{G}),
\end{equation}
and say that $a$ is \emph{valid} on $\mathcal{G}$, or achieves recourse on $\mathcal{G}$, when $\Delta(a;\mathcal{G})\geq0$.  The expectation and variance in $\Delta(a;\mathcal{G})$ are intractable for a general classifier $h$, so we estimate them by Monte Carlo as in~\citet{karimiAlgorithmicRecourseImperfect2020}; we write $\hat{\Delta}_M$ for the estimate from $M$ samples.

\paragraph{Affordability: minimizing the cost of the action.}
In the GP posterior experiments we define the cost \rev{$\costF(a)$} as the Euclidean distance between the factual values of the intervened features and their intervention values, $\costF(a)=\lVert\btheta-\xF_{\mathcal{I}}\rVert_2$ for $a=do(\XI=\btheta)$; the supplied-equation controls and experiments using linear regression fitted by ordinary least squares use a weighted $L_1$ cost (\cref{app:scms}).  The cost depends only on the action, so it is the same under every graph.

\paragraph{Robustness: remaining valid across plausible graphs.}
Since the true underlying graph is not known, we let $\GG$ denote a supplied family of candidate DAGs. Our main structural-ambiguity setting uses the MEC of a given \rev{completed partially directed acyclic graph (CPDAG), which represents the MEC by showing arrows only along edges that have the same direction in every DAG of the class (\cref{app:mec}). In practice, the CPDAG would be estimated from observational data by causal discovery~\citep{spirtes2000causation, chickering2002optimal}; this is outside the scope of our work.} For each $\mathcal{G}\in\GG$, we fit a GP-SCM $\mathcal{M}_{\mathcal{G}}$ \eqref{eq:gp-scm} on the same training data, which gives one counterfactual random variable $\XCF(a;\mathcal{G})$, and one margin $\Delta(a;\mathcal{G})$, per candidate graph.  The threshold also depends on the graph, because each GP-SCM implies a different counterfactual variance.  An action that is valid and cheap under one graph can fail under another (\cref{fig:motivation}), so we extend \eqref{eq:prob-recourse} by requiring its constraint on every candidate.  For each factual $\xF$, we seek
\begin{equation}\label{eq:objective}
a^{\texttt{F}}\in
 \operatorname*{arg\,min}_{a=do(\XI=\btheta)\in\AF} \costF(a) \quad\text{subject to}\quad
 \Delta(a;\mathcal{G})\geq0 \quad \forall\,\mathcal{G}\in\GG.
\end{equation}
We call the action \emph{fully robust} when it is valid on every $\mathcal{G}\in\GG$, so \eqref{eq:objective} seeks the cheapest fully robust action.

Requiring validity on every graph can, however, leave some individuals with no action at all, so \cref{sec:alg2} relaxes the requirement to a specified fraction of graph weight.  We evaluate both \rev{algorithms} by \emph{coverage}, the fraction of factual instances for which an action is returned, and \emph{true validity}, the fraction of returned actions satisfying $h(\xCF(a;\mathcal G^\star))\geq0.5$ under the generating SCM, which is known in our simulated experiments.

\subsection{Algorithm 1: Worst-Case Recourse}\label{sec:alg1}

\Cref{alg:worst-case} searches for the lowest cost action that would obtain a favorable outcome on every graph in $\GG$.  The full procedure is illustrated in \cref{fig:pipeline}.

For each graph $\mathcal{G}$, we score the outcome by passing its margin through a sigmoid:
\begin{equation}\label{eq:validity}
V_{\mathcal{G},\kappa}(a)=\sigma\big(\kappa\,\Delta(a;\mathcal{G})\big),
\end{equation}
where $V_{\mathcal{G},\kappa}(a)\to1$ indicates a favorable outcome for an action on the graph $\mathcal{G}$, and $V_{\mathcal{G},\kappa}(a)\geq0.5$ exactly when $\Delta(a;\mathcal{G})\geq0$.  The parameter $\kappa$ controls the `sharpness' of this \emph{soft validity}.  As $\kappa\to\infty$, $V_{\mathcal{G},\kappa}$ approaches a step function at $\Delta=0$: we only need the expected output to clear the threshold, and a larger margin brings little additional benefit.  Smaller values of $\kappa$ give a smoother score that rewards actions pushing the expected output further past the threshold.

We evaluate performance across graphs with a smooth lower envelope controlled by $\beta$:

\begin{equation}\label{eq:softmin}
\tilde{V}_{\beta}(a)=-\frac1\beta\log\sum_{\mathcal{G}\in\GG}
       \exp\big(-\beta\, V_{\mathcal{G},\kappa}(a)\big).
\end{equation}

This means that graphs that produce worse outcomes have a higher impact on the loss.  As $\beta \to \infty$, the validity of an action is determined only by the outcome on the worst graph.  We encourage the overall validity of the action, $\tilde{V}_{\beta}$ to surpass some prespecified threshold $\tau$.  In practice, $\Delta$ in \eqref{eq:validity} is replaced by its Monte Carlo estimate $\hat{\Delta}_M$.

For each possible intervention set $\mathcal{I}$, we search for low-cost intervention values $\btheta$ by using gradient descent with the following loss function:
\begin{equation}\label{eq:loss}
\mathcal{L}(\btheta)=
\costF(a)
+\lambda\big[\tau-\tilde{V}_{\beta}(a)\big]_+,
\qquad a=do(\XI=\btheta).
\end{equation}

We evaluate every candidate action on every graph using
$M'\geq M$ fresh counterfactual samples, independent of
the gradient samples. An action passes verification when
$\hat\Delta_{M'}(a;\mathcal G)\geq0$ for every candidate graph.
We return the cheapest passing action across iterates,
target sets and initializations, and assess it with an
additional independent post-selection evaluation.

Each initialization takes $T$ gradient steps of size $\eta$. $\Pi_{\Theta_{\mathcal{I}}}$ projects onto feasible intervention values for $\mathcal{I}$.  We compute $\nabla_{\btheta}\mathcal{L}$ through reparametrized samples, including the variance gradient when $\gamma_{\textsc{lcb}}\neq0$~\citep[App.~F]{karimiAlgorithmicRecourseImperfect2020}. The returned $a^\star$ approximates $a^{\texttt{F}}$.

\begin{figure}[t]
    \centering
    \scalebox{1.0}{\begin{minipage}{\linewidth}
    \centering
    \begin{minipage}[b]{.40\linewidth}
        \centering
        \raisebox{0mm}[\height][\depth]{% Explicit outer node boundary: incident edges stop at the OUTER ring.
\begin{tikzpicture}[x=1cm,y=1cm,
 v/.style={circle,draw,line width=.4pt,minimum size=7mm,inner sep=0pt,outer sep=.2pt,fill=white},
 every node/.style={font=\small}]
\node[v,minimum size=8.2mm] (x1) at (0,0) {$X_1$};
\draw[line width=.4pt] (x1.center) circle[radius=3.4mm];
\node[v] (x2) at (-.85,1.05) {$X_2$};
\node[v] (x3) at (.85,1.05) {$X_3$};
\node[v] (x6) at (1.05,-.85) {$X_6$};
\node[v] (x20) at (-1.05,-.85) {$X_{20}$};
\draw[line width=.4pt] (x1)--(x2) (x1)--(x3) (x1)--(x6) (x1)--(x20);
\node at (0,-1.35) {$\cdots$};
\node[align=center,font=\small] at (0,-1.95)
{20-node star CPDAG\\Only $X_1$ is actionable};
\end{tikzpicture}}\\[3pt]
        {\selectfont (a) Fixed causal structure}
    \end{minipage}\qquad
    \begin{minipage}[b]{.50\linewidth}
        \centering
        \includegraphics[width=\linewidth]{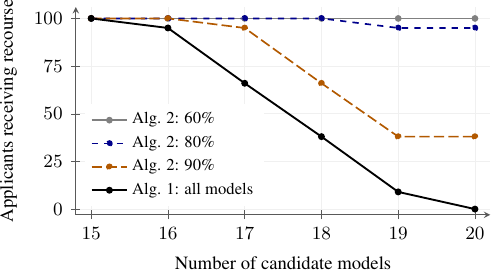}\\[3pt]
        {\selectfont (b) Expanding the candidate family}
    \end{minipage}
    \end{minipage}}
    
    \caption{
        (a) The fixed 20-node star CPDAG, shown schematically.
        Omitted leaves are also adjacent only to $X_1$.
        The double circle marks the only actionable variable.
        Each compatible DAG is oriented away from one root.
        (b) Returned actions as the candidate family expands;
        counts agree with exact feasibility.
        All methods serve 100 factual instances at prefixes 1--15.
        Only the full 20-model family is the complete MEC.
        Percentages denote required graph weight, uniform within
        each prefix.
    }
    \label{fig:graded-motivation}

\end{figure}
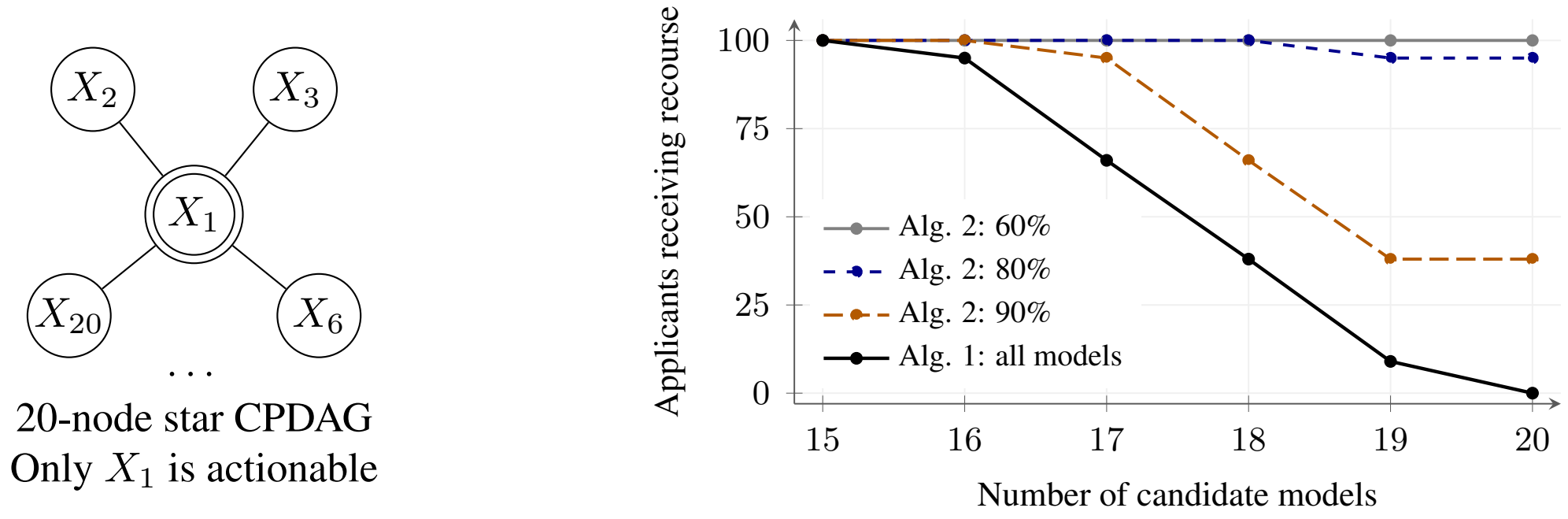

\paragraph{Why relax full robustness?}
Full robustness can rule out every permitted action even when recourse exists under some, possibly the generating, graph.  \rev{For example, if the only actionable feature shares an undirected CPDAG edge with the feature the classifier depends on, then any DAG orienting that edge into the actionable feature makes every action fail, even when the opposite orientation admits recourse (\cref{prop:impossibility} in \cref{app:impossibility}).}  The same effect can build up as the candidate family grows.  Our controlled example generates data from a 20-node star DAG whose CPDAG (\cref{fig:graded-motivation}a) admits 20 DAGs. We run \cref{alg:worst-case} on 100 factual instances as the candidate family expands from one DAG to the full MEC.  The classifier and candidate order are deliberately chosen to introduce increasingly restrictive constraints. Appendix~\ref{app:graded-motivation} gives the construction.  \cref{alg:worst-case} finds recourse for every feasible factual instance, but its coverage falls as additional models restrict the set of common permitted actions (\cref{fig:graded-motivation}b), so the decline reflects infeasibility rather than missed feasible solutions.  This motivates relaxing full robustness to validity under a specified fraction of graph weight.  With the full 20-model family, \Cref{alg:weighted} serves 100, 95 and 38 factual instances at required weights of 60\%, 80\% and 90\%, sacrificing validity under some compatible models.

\subsection{Algorithm 2: Graph-Weighted Recourse}\label{sec:alg2}

\Cref{alg:weighted} relaxes full robustness, requiring the action to pass the per-graph criterion under a specified total graph weight rather than every candidate graph.  Observational conditional independences alone do not identify which DAG within an MEC is correct, so domain knowledge or additional modeling assumptions can motivate these weights~\citep{meek1995causal}. Without a preferred graph, we use uniform weights.  We denote the weights on the candidate DAGs as $\pi=\{\pi_{\mathcal{G}}\}_{\mathcal{G}\in\GG}$. We require $\pi_{\mathcal G}\geq0$ and $\sum_{\mathcal G\in\GG}\pi_{\mathcal G}=1$.

\cref{alg:weighted} differs from the first in three places: the recourse constraint, the validity \rev{term of the loss \eqref{eq:loss}}, and the verification step\rev{, shown in blue in \cref{alg:worst-case,alg:weighted}; the latter two are also shaded in \cref{fig:pipeline}.}

\paragraph{Weighted objective.}  Using these graph weights $\pi$, we define the \emph{robustness} \rev{\emph{score}} of an action as the total weight of the graphs on which it is valid:
\begin{equation}\label{eq:robustness}
R_\pi(a)=\sum_{\mathcal{G}\in\GG}\pi_{\mathcal{G}}\,\mathbf{1}\{\Delta(a;\mathcal{G})\geq0\}.
\end{equation}
We then replace the constraint in \eqref{eq:objective} with $R_\pi(a)\geq1-\varepsilon$, where $\varepsilon\in[0,1)$ is the total graph weight on which the action is permitted to fail.  We call an action satisfying this constraint \emph{$(1-\varepsilon)$-robust}.  When $\pi$ is read as a distribution over graphs, this is a chance constraint~\citep{charnes1959chance, nemirovski2006convex}.  Note that $R_\pi$ averages \textit{whether} recourse is achieved on each graph, not the classifier output itself.  Averaging outputs would let a large margin on one graph offset failures on others.  With uniform weights $\pi_{\mathcal{G}}=1/|\GG|$, $\varepsilon$ is simply the fraction of graphs on which recourse may fail.  If all graph weights are positive, $\varepsilon=0$ recovers \cref{alg:worst-case}'s objective.

\paragraph{Loss function.}  Since we no longer require recourse on the worst graph, we replace the soft minimum validity $\tilde{V}_\beta$ in the loss function with a weighted average of the soft validities.  At each update, we compute one fresh Monte Carlo score for
every candidate graph and sample $K$ graph indices to form
the search aggregate, reusing scores for repeated indices.  For independent indices $\mathcal{G}_1,\ldots,\mathcal{G}_K\sim\pi$, we use
\begin{equation}\label{eq:weighted-loss}
\mathcal{L}(\btheta)=
\costF(a)
+\lambda\big[\tau-\hat{V}_{\pi}(a)\big]_+,
\qquad
\hat{V}_{\pi}(a)=\frac1K\sum_{k=1}^K V_{\mathcal{G}_k,\kappa}(a),
\end{equation}
Here $\hat V_\pi$, conditional on the computed Monte Carlo scores, is an unbiased estimate of the weighted average $\sum_{\mathcal{G}\in\GG}\pi_{\mathcal{G}}V_{\mathcal{G},\kappa}(a)$.  Graphs with greater weight are sampled more often and therefore have more influence on the gradient. Graphs on which the action fails are no longer emphasized as in \eqref{eq:softmin}.  $\tau$ is the smooth-score target guiding gradient steps, while $1-\varepsilon$ is the required graph weight used in verification.

\paragraph{Verification.}  \rev{As in \cref{alg:worst-case}, each} candidate is verified on every $\mathcal{G}\in\GG$ with $M'$ fresh samples, and accepted if
\begin{equation}\label{eq:mc-robustness}
\hat{R}_\pi(a)=\sum_{\mathcal{G}\in\GG}\pi_{\mathcal{G}}\,\mathbf{1}\{\hat{\Delta}_{M'}(a;\mathcal{G})\geq0\}\geq1-\varepsilon.
\end{equation}
The cost of verification therefore still grows with $|\GG|$.

% The passing criterion is relative to $\pi$ and the fitted models: if the weights do not reflect which graph is true, an action may satisfy $\hat{R}_\pi(a)\geq1-\varepsilon$ and still fail under the true graph.

% Side-by-side comparison; content and reference labels are unchanged.
\begin{figure}[t!]
\centering
\begingroup
\newcommand{\algdiff}[1]{\textcolor{blue}{#1}}
\SetAlFnt{\footnotesize}
\SetAlCapFnt{\footnotesize}
\SetAlCapNameFnt{\footnotesize}
\begin{minipage}[t]{0.485\linewidth}
\vspace{0pt}
\begin{algorithm2e}[H]
\caption{Worst-case search}
\label{alg:worst-case}
\KwIn{Factual $\xF$; classifier $h$; GP-SCMs $\{\mathcal{M}_{\mathcal{G}}\}_{\mathcal{G}\in\GG}$; target sets $\mathfrak{I}$; feasible values $\Theta_{\mathcal{I}}$; cost $\costF$; parameters $\eta,\lambda,\kappa,\beta,\tau,\gamma_{\textsc{lcb}},T,M,M'$.}
\vspace{\baselineskip}
\KwOut{An action $a^\star$, or ``no action found''.}
Set $a^\star$ to empty and its cost to $+\infty$\;
\ForEach{$\mathcal{I}\in\mathfrak{I}$ and initialization $\btheta\in\Theta_{\mathcal{I}}$}{
 \For{$t=0,\ldots,T$}{
  Form $a=do(\XI=\btheta)$ and compute $\hat{\Delta}_{M'}(a;\mathcal{G})$ on fresh samples for every $\mathcal{G}\in\GG$\;
  \If{\algdiff{$\hat{\Delta}_{M'}(a;\mathcal{G})\geq0$ for every $\mathcal{G}\in\GG$}}{
   Keep $a$ if cheaper than $a^\star$\;
  }
  \If{$t<T$}{
   \vspace{\dimexpr\baselineskip+1pt\relax}
   Compute $\hat{\Delta}_M(a;\mathcal{G})$, $V_{\mathcal{G},\kappa}(a)$, and \algdiff{$\tilde{V}_\beta(a)$}, then $\mathcal{L}(\btheta)$ from \algdiff{\eqref{eq:loss}}\;
   Update $\btheta\leftarrow\Pi_{\Theta_{\mathcal{I}}}(\btheta-\eta\nabla_{\btheta}\mathcal{L})$\;
  }
 }
}
Return $a^\star$ if found; else ``no action found''\;
\end{algorithm2e}
\end{minipage}\hfill
\begin{minipage}[t]{0.485\linewidth}
\vspace{0pt}
\begin{algorithm2e}[H]
\caption{Graph-weighted search}
\label{alg:weighted}
\KwIn{Factual $\xF$; classifier $h$; GP-SCMs $\{\mathcal{M}_{\mathcal{G}}\}_{\mathcal{G}\in\GG}$; graph weights $\pi$; permitted failure weight $\varepsilon$; target sets $\mathfrak{I}$; feasible values $\Theta_{\mathcal{I}}$; cost $\costF$; parameters $\eta,\lambda,\kappa,\tau,\gamma_{\textsc{lcb}},T,K,M,M'$.}
\KwOut{An action $a^\star$, or ``no action found''.}
Set $a^\star$ to empty and its cost to $+\infty$\;
\ForEach{$\mathcal{I}\in\mathfrak{I}$ and initialization $\btheta\in\Theta_{\mathcal{I}}$}{
 \For{$t=0,\ldots,T$}{
  Form $a=do(\XI=\btheta)$ and compute $\hat{\Delta}_{M'}(a;\mathcal{G})$ on fresh samples for every $\mathcal{G}\in\GG$\;
  \If{\algdiff{$\hat{R}_\pi(a)\geq1-\varepsilon$}}{
   Keep $a$ if cheaper than $a^\star$\;
  }
  \If{$t<T$}{
   \algdiff{Draw $\mathcal{G}_1,\ldots,\mathcal{G}_K\sim\pi$ independently}\;
   Compute $\hat{\Delta}_M(a;\mathcal{G}_k)$, $V_{\mathcal{G}_k,\kappa}(a)$, and \algdiff{$\hat{V}_\pi(a)$}, then $\mathcal{L}(\btheta)$ from \algdiff{\eqref{eq:weighted-loss}}\;
   Update $\btheta\leftarrow\Pi_{\Theta_{\mathcal{I}}}(\btheta-\eta\nabla_{\btheta}\mathcal{L})$\;
  }
 }
}
Return $a^\star$ if found; else ``no action found''\;
\end{algorithm2e}
\end{minipage}
\endgroup
\end{figure}

% \subsection{Algorithm 2 Results}

% \section{Results}

% Report number of valid actions returned for each method

% Report how action cost changes, how computational cost (runtime) changes

% Discuss worst case, best case, average case...

\section{Experiments}\label{sec:experiments}

\subsection{Setup and evaluation}\label{sec:setup}
\begin{figure}[t]
\centering
\scalebox{1.00947}{% Native TikZ: inherits the manuscript's text and math fonts.
\begin{tikzpicture}[x=1cm,y=1cm,>=Latex,
 v/.style={circle,draw,thin,minimum size=7mm,inner sep=1pt},
 every node/.style={font=\small},every path/.style={line width=.4pt}]
\node at (1,1.9) {(a) Three variables};
\node[v] (a1) at (1,1) {$X_1$};
\node[v] (a2) at (.25,0) {$X_2$};
\node[v] (a3) at (1.75,0) {$X_3$};
\draw (a1)--(a2)--(a3)--(a1);
\node at (1,-.65) {CPDAG: six DAGs};
\node[v] (b1) at (1,-1.6) {$X_1$};
\node[v] (b2) at (.25,-2.6) {$X_2$};
\node[v] (b3) at (1.75,-2.6) {$X_3$};
\draw[->] (b1)--(b2); \draw[->] (b1)--(b3); \draw[->] (b2)--(b3);
\node at (1,-3.25) {Generating DAG};
\node at (5,1.9) {(b) Linear chains};
\node[v] (c1) at (3.5,.5) {$X_1$};
\node[v] (c2) at (4.5,.5) {$X_2$};
\node (cc) at (5.5,.5) {$\cdots$};
\node[v] (cd) at (6.5,.5) {$X_d$};
\draw (c1)--(c2)--(cc)--(cd);
\node at (5,-.65) {CPDAG: $d$ DAGs};
\node[v] (d1) at (3.5,-2.1) {$X_1$};
\node[v] (d2) at (4.5,-2.1) {$X_2$};
\node (dd) at (5.5,-2.1) {$\cdots$};
\node[v] (de) at (6.5,-2.1) {$X_d$};
\draw[->] (d1)--(d2); \draw[->] (d2)--(dd); \draw[->] (dd)--(de);
\node at (5,-3.25) {Generating DAG};
\begin{scope}[xshift=1cm]
\node at (9,1.9) {(c) Nonlinear ANM};
\node[v] (e1) at (9,1) {$X_1$};
\node[v] (e2) at (8.25,0) {$X_2$};
\node[v] (e3) at (9.75,0) {$X_3$};
\draw[->] (e1)--(e2); \draw[->] (e1)--(e3); \draw[->] (e2)--(e3);
\node at (9,-.65) {One supplied DAG};
\node[align=left,font=\footnotesize] at (9,-2.1) {$X_1=U_1$\\[3pt]$X_2=-1+3\sigma(2X_1)+U_2$\\[3pt]$X_3=-.05X_1+.25X_2^2+U_3$};
\end{scope}
\end{tikzpicture}}
\caption{Causal structures. (a) Three-variable CPDAG (six DAGs) for linear, nonlinear-classifier and transformed nonlinear-SCM settings. (b) Chain CPDAG: $d$ DAGs, one directed away from each root. (c) Supplied DAG: nonlinear additive-noise model with independent exogenous noises $U_1,U_2,U_3$. Targets: up to two of $\{X_1,X_3\}$ in (a) or $\{X_1,X_d\}$ in (b), or any one variable in (c). Lower DAGs in (a)--(b) identify the truth for readers only.}
\label{fig:experimental-structures}
\end{figure}
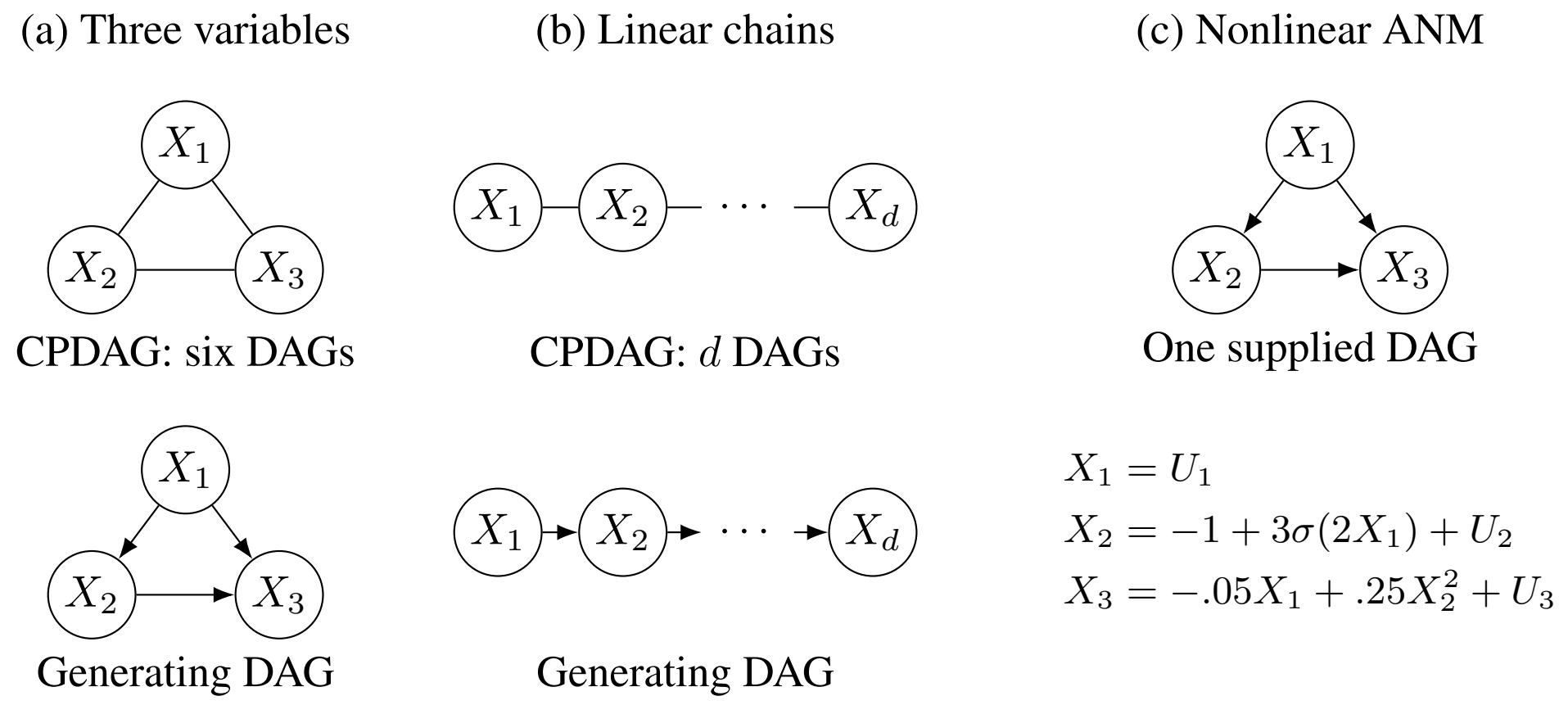

We evaluate fitted linear regression structural equations as a baseline with only graph uncertainty and use GP-SCMs in the rest of our experiments.
Both algorithms share factual instances, classifiers, action restrictions, and fitted mechanisms within each setting.
Observational data is used to fit the structural equations, while the generating SCM
is only used as ground truth to check validity. \Cref{fig:experimental-structures} summarizes the small linear and nonlinear settings, larger linear chains, and supplied-graph nonlinear companion. We also consider a semi-synthetic loan
problem with Option 1 (a four-DAG MEC) and Option 2 (a broader eight-DAG stress test)
(\cref{app:loan-uncertain}). 

Our SCMs are fitted using 128 observed data points. Recourse is determined for 1,000 factual instances obtained from a total of 10 seeds.  \cref{alg:weighted} uses uniform graph weights, with required \rev{graph weight}
$1-\varepsilon=0.8$ for synthetic settings and $0.5$ for the
main loan comparison. Costs follow \cref{sec:methodology};
fitting and search settings are detailed in
\cref{app:fitted-protocol} and \cref{tab:hyperparameters}. Supplied-equation controls are reported in Appendix~\ref{app:population-controls}. \rev{Appendix~\ref{app:single-dag} compares both algorithms with recourse on one uniformly selected DAG; Appendix~\ref{app:optimization-gaps} reports exact-reference optimization gaps for the population controls.}

We report coverage, true validity, cost, and runtime.
Tables show equal-weight seed means and sample standard
deviations. True validity and cost condition on returned
actions and omit seeds with no returns; coverage includes
all seeds. New-sample validity (passed/returned) pools counts across seeds after rechecking each returned action on 512 fresh counterfactual samples from the fitted models. The factual instances and actions stay fixed; this check differs from success under the generating SCM.
Paired costs and pooled counts appear in Appendices~\ref{app:fitted-paired} and~\ref{app:fitted-details}, respectively. Code and experiment configurations are available at
\url{https://anonymous.4open.science/r/robust-recourse-9E6C/}.

\subsection{Effect of mechanism fitting}\label{sec:ols-results}\label{sec:gp-results}
\rev{We fit each candidate graph's mechanisms either by linear regression, whose counterfactuals are deterministic but can approximate nonlinear mechanisms poorly (\cref{tab:ols-results}), or by GP posteriors, which propagate uncertainty in the equations (\cref{tab:gp-lcb}). Under both fits, coverage is high in small settings but falls on larger chains. The GP posterior raises true validity in most settings, most clearly for \cref{alg:weighted} and on the supplied-graph ANM (64.8\% to 99.4\%). On 20-node chains, \cref{alg:weighted} returns more actions than \cref{alg:worst-case} but has lower true validity under either fit: with uniform weights over 20 DAGs, the 0.8 requirement permits four graphs to fail, including the generating graph. Appendix~\ref{app:graded-motivation} distinguishes structural infeasibility from search failure in a controlled study; fitted no-action cases are not proved infeasible.}

\begin{table}[tbp]
\centering
\caption{Fitted linear regression baseline. Mean $\pm$ sample SD across ten seeds, 100 factual instances per seed. \cref{alg:weighted} requires graph weight 0.8. Cost is weighted $L_1$ among returned actions; comparisons are not paired. Conditional statistics omit no-return seeds.}
\label{tab:ols-results}
\footnotesize
\setlength{\tabcolsep}{3pt}
\renewcommand{\arraystretch}{1.08}
\begin{lrbox}{\revisiontable}
\begin{tabular}{@{}lrrrrrr@{}}
\toprule
& \multicolumn{3}{c}{\cref{alg:worst-case}}
& \multicolumn{3}{c}{\cref{alg:weighted}} \\
\cmidrule(lr){2-4}\cmidrule(l){5-7}
Setting
& Coverage (\%)
& True validity (\%)
& Cost
& Coverage (\%)
& True validity (\%)
& Cost \\
\midrule
Linear, 3 nodes & { \ms{92.0}{4.3}} & { \ms{90.1}{21.8}} & { \ms{4.96}{0.40}} & { \ms{94.2}{4.2}} & { \ms{72.5}{26.2}} & { \ms{4.78}{0.47}} \\
Nonlinear classifier & { \ms{92.1}{3.2}} & { \ms{91.0}{18.2}} & { \ms{4.99}{0.48}} & { \ms{95.7}{3.3}} & { \ms{71.1}{24.2}} & { \ms{4.91}{0.51}} \\
Nonlinear SCM & { \ms{81.1}{6.4}} & { \ms{92.1}{17.0}} & { \ms{5.15}{0.45}} & { \ms{84.9}{7.7}} & { \ms{75.1}{24.9}} & { \ms{5.21}{0.56}} \\
Supplied-graph ANM & { \ms{99.5}{0.5}} & { \ms{64.8}{5.7}} & { \ms{0.19}{0.01}} & { \ms{99.5}{0.5}} & { \ms{64.8}{5.7}} & { \ms{0.19}{0.01}} \\
\midrule
Linear, 5 nodes & { \ms{82.3}{11.8}} & { \ms{85.0}{27.6}} & { \ms{4.77}{1.84}} & { \ms{90.2}{5.7}} & { \ms{45.5}{48.2}} & { \ms{3.50}{0.85}} \\
Linear, 10 nodes & { \ms{61.2}{14.7}} & { \ms{83.2}{28.6}} & { \ms{5.37}{1.39}} & { \ms{69.7}{10.2}} & { \ms{42.9}{49.8}} & { \ms{4.53}{0.80}} \\
Linear, 20 nodes & { \ms{35.8}{12.1}} & { \ms{78.4}{41.6}} & { \ms{6.27}{1.25}} & { \ms{43.9}{9.3}} & { \ms{41.0}{50.9}} & { \ms{6.06}{1.02}} \\
\bottomrule
\end{tabular}\end{lrbox}
\ifdim\wd\revisiontable>\linewidth\resizebox{\linewidth}{!}{\usebox{\revisiontable}}\else\usebox{\revisiontable}\fi
\end{table}

\begin{table}[tbp]
\centering\footnotesize
\caption{Main GP posterior results: ten seeds, 100 factual instances per seed. Entries are seed mean $\pm$ sample SD, except new-sample validity: pooled passed/returned counts and percentages from an independent 512-draw check. \cref{alg:weighted} requires graph weight 0.8. Euclidean costs condition on returns and are unpaired; conditional statistics omit no-return seeds.}
\label{tab:gp-lcb}
\setlength{\tabcolsep}{3pt}\renewcommand{\arraystretch}{1.08}
\begin{lrbox}{\revisiontable}
\begin{tabular}{@{}lrrrrrrrr@{}}
\toprule
& \multicolumn{4}{c}{\cref{alg:worst-case}} & \multicolumn{4}{c}{\cref{alg:weighted}} \\
\cmidrule(lr){2-5}\cmidrule(l){6-9}
Setting & \shortstack[r]{Coverage\\(\%)} & \shortstack[r]{True\\validity (\%)} & Cost & \shortstack[r]{New-sample validity\\passed/returned (\%)} & \shortstack[r]{Coverage\\(\%)} & \shortstack[r]{True\\validity (\%)} & Cost & \shortstack[r]{New-sample validity\\passed/returned (\%)} \\
\midrule
Linear, 3 nodes & { \ms{90.6}{3.4}} & { \ms{98.9}{1.7}} & { \ms{1.33}{0.10}} & { 883/906 (97.5\%)} & { \ms{92.5}{4.6}} & { \ms{95.4}{4.5}} & { \ms{1.32}{0.09}} & { 910/925 (98.4\%)} \\
Nonlinear classifier & { \ms{90.3}{3.1}} & { \ms{99.4}{1.5}} & { \ms{1.34}{0.10}} & { 879/903 (97.3\%)} & { \ms{93.3}{4.2}} & { \ms{96.2}{2.9}} & { \ms{1.34}{0.09}} & { 924/933 (99.0\%)} \\
Nonlinear SCM & { \ms{79.2}{5.7}} & { \ms{97.8}{2.9}} & { \ms{1.38}{0.14}} & { 758/792 (95.7\%)} & { \ms{82.2}{6.3}} & { \ms{93.7}{4.3}} & { \ms{1.38}{0.11}} & { 772/822 (93.9\%)} \\
Supplied-graph ANM & { \ms{89.3}{3.0}} & { \ms{99.4}{0.9}} & { \ms{1.99}{0.17}} & { 866/893 (97.0\%)} & { \ms{89.3}{3.0}} & { \ms{99.4}{0.9}} & { \ms{1.99}{0.17}} & { 866/893 (97.0\%)} \\
Linear, 10 nodes & { \ms{54.8}{16.5}} & { \ms{93.0}{10.7}} & { \ms{2.01}{0.45}} & { 433/548 (79.0\%)} & { \ms{68.8}{12.0}} & { \ms{51.6}{44.4}} & { \ms{1.75}{0.16}} & { 688/688 (100.0\%)} \\
Linear, 20 nodes & { \ms{34.1}{11.1}} & { \ms{76.8}{34.8}} & { \ms{2.23}{0.22}} & { 229/341 (67.2\%)} & { \ms{43.6}{9.1}} & { \ms{43.4}{49.3}} & { \ms{2.14}{0.23}} & { 436/436 (100.0\%)} \\
\bottomrule\end{tabular}\end{lrbox}
\ifdim\wd\revisiontable>\linewidth\resizebox{\linewidth}{!}{\usebox{\revisiontable}}\else\usebox{\revisiontable}\fi
\end{table}

\subsection{Semi-synthetic loan results}\label{sec:loan-gp-uncertain}
The semi-synthetic loan study tests recourse with uncertain causal directions and restricted actions. Option 1 is a four-DAG MEC: education--income and loan amount--duration can each point in either direction, with all other edges fixed. Option 2 is an eight-DAG family that additionally allows either direction for income--savings.
The extra graphs are outside the generating MEC and test sensitivity to additional structural assumptions. Both families include the generating graph with uniform weights.
Fitting graph-specific GP mechanisms to the same observational data introduces mechanism-estimation error alongside graph uncertainty.

Each action increases one permitted feature. We compare
allowing and excluding direct savings interventions;
education, loan amount, and income remain actionable.
\cref{alg:worst-case} requires every candidate to pass the posterior
score criterion. For \cref{alg:weighted}, we use graph weight 0.5
to provide a genuine relaxation in both families: two of
four or four of eight candidates must pass. At 0.8 and 0.9, all four MEC candidates must pass,
as required by \cref{alg:worst-case}.

\cref{tab:loan-main} reports coverage, true validity, cost, and new-sample validity (passed/returned). Appendices~\ref{app:loan-uncertain} and~\ref{app:loan-thresholds} detail graph structures, action bounds, and fitting, and the linear regression baseline and posterior threshold sweep, respectively.

\begin{table}[tbp]
\centering\footnotesize
\caption{Loan GP posterior results: ten seeds, 100 applicants per seed for each family/action setting. \cref{alg:weighted} requires graph weight 0.5. Entries are seed mean $\pm$ sample SD; costs are Euclidean. Conditional statistics omit no-return seeds. New-sample validity reports pooled passed/returned counts and percentages from independent 512-draw evaluations. Option 1: four-DAG MEC; Option 2: eight-DAG candidate family.}
\label{tab:loan-main}
\setlength{\tabcolsep}{3pt}
\renewcommand{\arraystretch}{1.08}

\newcommand{\loanrows}{%
\toprule
& & \multicolumn{4}{c}{\cref{alg:worst-case}}
  & \multicolumn{4}{c}{\cref{alg:weighted}} \\
\cmidrule(lr){3-6}\cmidrule(l){7-10}
\shortstack[l]{Direct savings\\allowed?}
& Graph family
& \shortstack[r]{Coverage\\(\%)}
& \shortstack[r]{True\\validity (\%)}
& Cost
& \shortstack[r]{New-sample validity\\passed/returned (\%)}
& \shortstack[r]{Coverage\\(\%)}
& \shortstack[r]{True\\validity (\%)}
& Cost
& \shortstack[r]{New-sample validity\\passed/returned (\%)} \\
\midrule
Yes & Option 1 & { \ms{93.7}{4.2}} & { \ms{98.7}{3.1}} & { \ms{6.60}{0.56}} & { 912/937 (97.3\%)} & { \ms{93.7}{4.2}} & { \ms{97.0}{4.2}} & { \ms{6.53}{0.60}} & { 890/937 (95.0\%)} \\
 & Option 2 & { \ms{92.2}{5.0}} & { \ms{99.9}{0.3}} & { \ms{7.06}{0.59}} & { 900/922 (97.6\%)} & { \ms{93.3}{4.4}} & { \ms{98.8}{1.9}} & { \ms{6.63}{0.56}} & { 919/933 (98.5\%)} \\
\midrule
No & Option 1 & { \ms{20.0}{16.5}} & { \ms{95.4}{9.3}} & { \ms{3.10}{1.24}} & { 149/200 (74.5\%)} & { \ms{22.5}{17.5}} & { \ms{84.6}{16.3}} & { \ms{2.95}{1.35}} & { 129/225 (57.3\%)} \\
 & Option 2 & { \ms{1.9}{3.8}} & { \ms{78.3}{43.9}} & { \ms{0.29}{0.08}} & { 15/19 (78.9\%)} & { \ms{20.2}{15.8}} & { \ms{89.3}{16.1}} & { \ms{3.03}{1.45}} & { 152/202 (75.2\%)} \\
\bottomrule
}

\begin{lrbox}{\revisiontable}
\begin{tabular}{@{}llrrrrrrrr@{}}
\loanrows
\end{tabular}
\end{lrbox}
\ifdim\wd\revisiontable>\linewidth
  \resizebox{\linewidth}{!}{\usebox{\revisiontable}}
\else
  \begin{tabular*}{\linewidth}
    {@{\extracolsep{\fill}}llrrrrrrrr@{}}
  \loanrows
  \end{tabular*}
\fi
\end{table}
\FloatBarrier

% With direct savings allowed, eight-graph coverage is 91/100 for Algorithm 1 and 92/100 for Algorithm 2. Without direct savings, it is 6/100 and 28/100, respectively; true successes are five and 26, and fresh posterior passes are 5/6 and 22/28. Relaxation can therefore increase coverage and the number of true successes in this loan setting, while the linear-chain study shows a loss of true validity. The outcome depends on decision-relevant responses, model fit and available actions. The large unpaired cost difference in the restricted loan row reflects different served populations: on six jointly served applicants, Algorithm 2 minus Algorithm 1 is $-0.07\pm0.11$ across contributing seeds.

\begin{figure}[t]

\centering
\includegraphics[width=\linewidth]{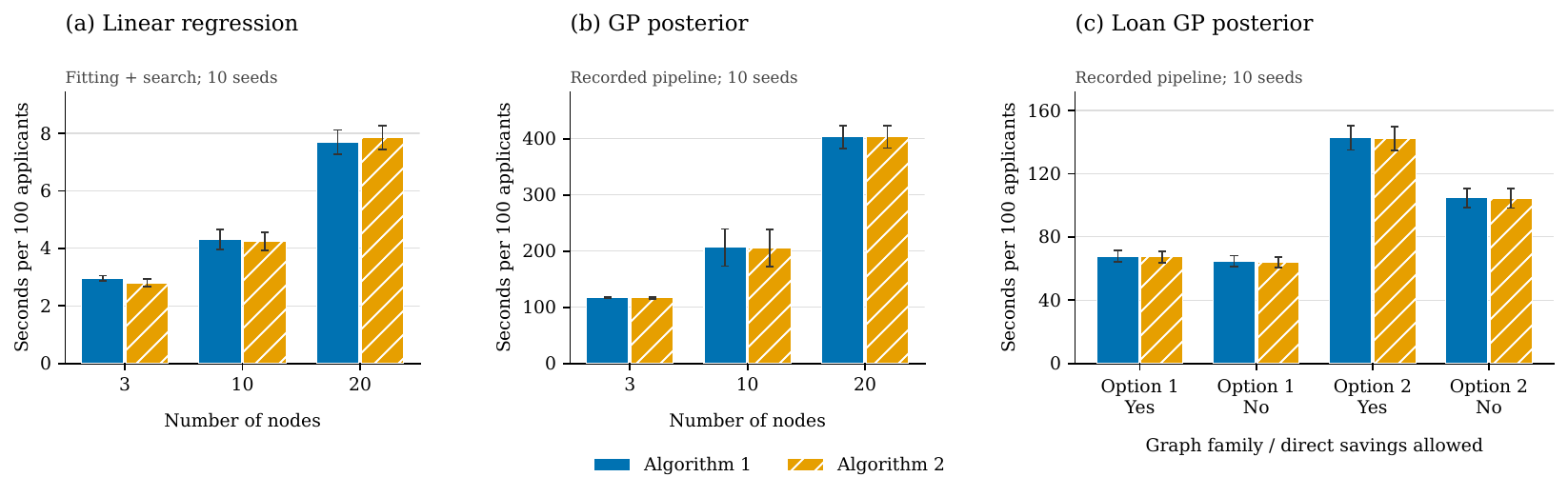}\par
\caption{Runtime per 100 factual instances across ten seeds (bars: means; error bars: sample SD). (a) Linear regression fitting and search; (b)--(c) GP posterior pipelines, counting shared setup and fitting once. \cref{alg:weighted} requires graph weight 0.8 (linear) or 0.5 (loan). Loan labels mark direct savings access; vertical scales differ.}
\label{fig:runtime}
\end{figure}

\paragraph{Computational cost.}
\Cref{fig:runtime} reports timings per 100 factual instances per fitted seed. Both algorithms evaluate every candidate graph and have similar runtimes within settings. Linear regression timings cover fitting and search; GP posterior timings also cover verification and final evaluation, charging shared setup and fitting once. Appendix~\ref{app:fitted-runtime} details timings, hardware and software, measurement scope, and reused measurements.

\section{Discussion}\label{sec:discussion}

\rev{The benefit of relaxation depends on the available actions and how the candidates respond to them, not only on the number of candidate graphs.} Relaxation is more useful when action restrictions make full robustness difficult to achieve. In the loan study under Option 2, allowing direct savings gives both algorithms high coverage\rev{, whereas excluding it} creates a larger coverage difference, and \cref{alg:weighted} also achieves more true successes (\cref{tab:loan-main}). \rev{When no common permitted action exists, as in the controlled study of Appendix~\ref{app:graded-motivation}, the loss of coverage under full robustness reflects infeasibility rather than search failure, and only relaxing the requirement can serve those individuals.}

\rev{Once the requirement is relaxed, graph weights} determine which causal possibilities receive emphasis. We use uniform DAG weights to give each candidate equal importance. Domain knowledge can motivate nonuniform weights, but misleading weights can increase returned coverage while reducing true validity, as shown in Appendix~\ref{app:weights}. The required graph weight provides a separate control over robustness (Appendix~\ref{app:loan-thresholds}). \rev{Because} these weights express the assumed causal possibilities, not a calibrated probability of real-world success, \rev{a reported action should state} its achieved graph weight and the candidates on which it fails. Weighted approval also differs from average classifier score; Appendix~\ref{app:criteria} compares these and additional acceptance conditions.

\rev{For the same reason,} passing the fitted-model criterion and succeeding under the generating SCM are distinct outcomes. In the 20-node posterior study, {436/436 \cref{alg:weighted} returns pass the fresh posterior check, but 184/436 succeed under the generating SCM} (\cref{tab:gp-lcb}). \rev{Robustness is therefore relative to the candidate family and its fitted mechanisms.} Both algorithms evaluate every candidate graph and have similar runtimes within each setting\rev{; runtime increases with chain size and is higher for loan Option 2 (\cref{fig:runtime}).}

\paragraph{Limitations.}
\rev{Robustness is relative to the candidate family and fitted mechanisms. We assume a supplied CPDAG and no latent confounding; discovery errors or hidden confounders can exclude the generating graph from $\GG$. Even within $\GG$, mechanism misspecification can cause accepted actions to fail, as linear regression does on the nonlinear ANM (\cref{tab:ols-results}). Gradient search may miss feasible actions; optimization gaps are measured only on linear population controls (Appendix~\ref{app:optimization-gaps}). Monte Carlo acceptance checks offer no calibrated statistical guarantees. Both algorithms fit and verify every DAG in the MEC, which can grow exponentially with undirected edges. Our larger experiments therefore use tractable chains; dense graphs remain untested. Evaluating true validity requires the generating SCM, restricting our evaluation to synthetic and semi-synthetic data.}

\section{Conclusion}\label{sec:conclusion}

Causal graph ambiguity affects robust recourse feasibility and cost. Individual candidates may admit recourse without a common action. \rev{In fitted studies, relaxation increased coverage (\cref{tab:ols-results,tab:gp-lcb}) and lowered paired costs (Appendix~\ref{app:fitted-paired}), but yielded more actions with fewer true successes on 20-node chains. True validity} depends on actions, graph family, and mechanism fit. Use \cref{alg:worst-case} when every candidate must pass the fitted-model criterion; use \cref{alg:weighted} when \rev{failing on} a specified \rev{fraction of graph weight} is acceptable. Future work could use neural intervention proposals, following CARMA~\citep{majumdar2024carma}, with separate robustness checks. Adapting local IDA~\citep{maathuis2009estimating} could reduce enumeration while preserving effects needed for verification.   In practice, CPDAG estimation can use constraint-based PC~\citep{spirtes2000causation} or score-based GES~\citep{chickering2002optimal} (see~\citet{glymour2019review}). Missing, extra, or misoriented edges from finite samples can exclude the truth from $\GG$.  To propagate discovery uncertainty, form $\GG$ from a posterior or bootstrap graph distribution~\citep{friedman2003being}, as in counterfactual fairness~\citep{valerioCounterfactualFairnessGraph2026}, with frequencies supplying \cref{alg:weighted}\textquotesingle s weights $\pi$.

\section*{Acknowledgments}
This work was supported by the Natural Sciences and Engineering Research Council of Canada (NSERC) through funding to Amir-Hossein Karimi and Mohammad Kohandel, with additional support from the O'Donovan family to Amir-Hossein Karimi.

\bibliography{yourbibfile}

\appendix

\section{Related Work}\label{sec:related}

\paragraph{Robust recourse.}  Prior work makes recourse robust to four sources of uncertainty, illustrated in \cref{fig:robustness-axes} in Appendix~\ref{app:robustness-axes}: errors in the recorded features~\citep{dominguez-olmedoAdversarialRobustnessCausal2022, weilbachEstimationCounterfactualInterventions2024}, imprecise execution of the action~\citep{pawelczyk2023probabilistically}, several equally good classifiers~\citep{pawelczyk2020multiplicity, dandlMultiObjectiveCounterfactualExplanations2020, russellEfficientSearchDiverse2019, black2022consistent}, and classifiers that are retrained over time~\citep{rawal2020wild, upadhyayRobustReliableAlgorithmic2021, nguyen2023distributionally, duellProvablyRobustBayesian2026, jiangFormalisingRobustnessCounterfactual2022, jiangProvablyRobustPlausible2024}.  Most of these methods change features directly and use no causal model.  The few that do use one~\citep{dominguez-olmedoAdversarialRobustnessCausal2022, weilbachEstimationCounterfactualInterventions2024} assume the causal graph is known.

\paragraph{Uncertainty in the causal graph.}  Uncertainty in the causal graph itself has also been studied in causal-effect estimation and fairness.  \citet{valerioCounterfactualFairnessGraph2026} assess counterfactual fairness~\citep{kusner2017counterfactual} across the graphs in an MEC by fitting a linear SCM to each graph.  \citet{singhCausalGuardConformalInference2026} and~\citet{chikaharaFairnessGraphUncertainty2026} assess interventional fairness without fitting structural equations:~\citet{singhCausalGuardConformalInference2026} weight each candidate graph by how well it fits the data, and~\citet{chikaharaFairnessGraphUncertainty2026} reduce the uncertainty by learning a coarser graph over clusters of variables~\citep{anandCausalDiscoveryClusters, anandCausalEffectIdentification2023}.

\begin{table}[t]
\centering
\small
\setlength{\tabcolsep}{3pt}
\renewcommand{\arraystretch}{1.15}
\newcommand{\cmark}{\ding{51}}\newcommand{\xmark}{\ding{55}}
\begin{tabular}{@{}>{\raggedright\arraybackslash}p{2.7cm}>{\raggedright\arraybackslash}p{2.2cm}>{\raggedright\arraybackslash}p{2.5cm}c>{\raggedright\arraybackslash}p{2.6cm}>{\raggedright\arraybackslash}p{2.8cm}@{}}
\toprule
\textbf{Method} & \textbf{Objective} & \textbf{Robust to} & \textbf{Causal} & \textbf{Structural equations} & \textbf{Causal graph} \\
\midrule
\textbf{Ours} & Recourse & Causal graph and structural equations & \cmark & GP-SCM fitted to each DAG & Supplied DAG family; main setting uses a CPDAG's MEC \\
\citet{valerioCounterfactualFairnessGraph2026} & Counterfactual fairness & Causal graph & \cmark & Linear SCM fitted to each DAG & DAGs from bootstrapped causal discovery \\
\citet{singhCausalGuardConformalInference2026} & Interventional fairness & Causal graph & \cmark & Not fitted & Candidate graphs weighted by fit to data \\
\citet{chikaharaFairnessGraphUncertainty2026} & Interventional fairness & Causal graph & \cmark & Not fitted & Coarser graph over clusters of variables \\
\midrule
\citet{karimiAlgorithmicRecourseImperfect2020} & Recourse & Structural equations & \cmark & GP-SCM fitted to the DAG & Known DAG \\
\citet{weilbachEstimationCounterfactualInterventions2024} & Recourse & Counterfactual ambiguity (equations and noise) & \cmark & Bayesian warped GP & Known DAG \\
\citet{dominguez-olmedoAdversarialRobustnessCausal2022} & Recourse & Recorded features & \cmark & Known SCM & Known DAG \\
\midrule
\citet{upadhyayRobustReliableAlgorithmic2021} & Recourse & Model changes & \xmark & -- & -- \\
\citet{duellProvablyRobustBayesian2026} & Counterfactual explanations & Model changes & \xmark & -- & -- \\
\citet{jiangFormalisingRobustnessCounterfactual2022} & Counterfactual explanations & Model changes & \xmark & -- & -- \\
\citet{jiangProvablyRobustPlausible2024} & Counterfactual explanations & Model changes & \xmark & -- & -- \\
\bottomrule
\end{tabular}
\caption{Comparison with related robust and causal methods.  \emph{Causal} indicates whether actions or counterfactuals are computed through a causal model; ``--'' means not applicable.  To our knowledge, ours is the only method that targets recourse while being robust to the causal graph.}
\label{tab:robust-methods}
\end{table}

\paragraph{Our contribution.}  \Cref{tab:robust-methods} compares our method with the most closely related work.  We bring uncertainty in the causal graph to algorithmic recourse.  This is a different task from fairness: rather than evaluating a fixed model, we search for an action that changes one individual's outcome, so we need that individual's counterfactual under each candidate graph.  Like~\citet{valerioCounterfactualFairnessGraph2026}, we fit an SCM to every graph in the MEC, but we use GP-SCMs instead of linear ones, which capture non-linear relationships and give a distribution over counterfactuals rather than a single value.  Our recommendations therefore account for uncertainty in the structural equations as well as in the graph.  Like~\citet{singhCausalGuardConformalInference2026}, we allow graphs to carry different weights, which our graph-weighted algorithm uses to relax full robustness.

\section{Markov equivalence and CPDAGs}\label{app:mec}

Causal graphs for real-world systems are seldom known, and a unique causal graph is not identifiable from most observational datasets.  Under causal sufficiency, acyclicity, and the causal Markov and faithfulness assumptions, population conditional independences identify a Markov equivalence class (MEC) of DAGs sharing those independences~\citep{pearl2016causal}.  \citet{verma1990equivalence} showed that two DAGs are Markov equivalent if and only if they have the same skeleton and the same v-structures.  All graphs in an MEC can be represented by a completed partially directed acyclic graph (CPDAG), also called an essential graph~\citep{andersson1997characterization, chickering2002learning}.  A CPDAG has directed edges only between nodes whose orientation is shared by every DAG in the MEC, and undirected edges otherwise; its orientations can be completed with the rules of~\citet{meek1995causal}.  In practice, a CPDAG is estimated from data by causal discovery, for example with PC~\citep{spirtes2000causation} or GES~\citep{chickering2002optimal}.  We take it as given (\cref{sec:background}) and enumerate the DAGs in its MEC to form the candidate family $\GG$.

\section{Impossibility of fully robust recourse}\label{app:impossibility}

\begin{observation}[Impossibility of fully robust recourse]\label[observation]{prop:impossibility}% [observation]: cleveref otherwise names it "Theorem" (shared counter)
Let $\GG$ be the MEC of a CPDAG containing an undirected edge $X_i - X_j$.  Suppose that $X_i$ is the only actionable feature, so that $\mathfrak{I}=\{\{i\}\}$, and let $\mathcal{G}^-\in\GG$ be a DAG containing $X_j\to X_i$.  Suppose that $h$ depends on $\mathbf{x}$ only through features other than $X_i$ that are not descendants of $X_i$ in $\mathcal{G}^-$ (for example, only through $x_j$).  Let $\xF$ satisfy $h(\xF)<0.5$.  Then, for every action $a=do(X_i=\theta)\in\AF$, $h(\xCF(a;\mathcal{G}^-))=h(\xF)<0.5$, so the fully robust problem \eqref{eq:objective} is infeasible.  This holds even if, under a DAG $\mathcal{G}^+\in\GG$ containing $X_i\to X_j$, some $a\in\AF$ achieves recourse.
\end{observation}

\begin{proof}
Because $X_i - X_j$ is undirected in the CPDAG, both orientations occur among the DAGs of the MEC, so $\mathcal{G}^-$ and $\mathcal{G}^+$ exist in $\GG$.  In $\mathcal{G}^-$, $X_j$ is a parent of $X_i$ and, by acyclicity, not a descendant of $X_i$.  Under $do(X_i=\theta)$, the counterfactual changes only $X_i$ and its descendants; every non-target non-descendant retains its factual value (abduction fixes the exogenous noise, and the structural equations of non-descendants do not involve $X_i$).  Hence $x_j$ and every other feature on which $h$ depends are unchanged, so $h(\xCF(a;\mathcal{G}^-))=h(\xF)<0.5$ for every $\theta$.  The constraint of \eqref{eq:objective} therefore fails on $\mathcal{G}^-$ for every feasible action, and no fully robust action exists.  Under $\mathcal{G}^+$, $X_j$ may change with $\theta$,
so recourse may exist under this candidate graph.  The same argument applies to the GP-SCM criterion \eqref{eq:thresh}: on $\mathcal{G}^-$ the counterfactual of $h$ is deterministic and equal to $h(\xF)$, so its variance is zero and the threshold is $0.5$.
\end{proof}

\section{SCMs used in experiments}\label{app:scms}
Throughout this section, $\mathcal N(\mu,v)$ denotes a Gaussian distribution with mean $\mu$ and variance $v$, and $\sigma$ is the sigmoid function. Candidate families are supplied. Population mechanisms isolate graph ambiguity; fitted linear regression and GP posterior mechanisms use the separate fitting and acceptance procedures described in the main text.

\subsection{Population model families}
\paragraph{Three-variable models.}
For each three-node seed $s$, let $\alpha_s=(6+(s\bmod10))/10$ and generate
\[
X_1=U_1,\quad X_2=-\alpha_sX_1+U_2,\quad
X_3=0.05\alpha_sX_1+0.25\alpha_sX_2+U_3,
\]
with independent centered Gaussian noises of variances $(1,0.8,1.2)$. The complete skeleton has six compatible DAGs. 

\paragraph{Linear chains.}
For $d\in\{5,10,20\}$,
\[
X_1=U_1,\quad X_j=b_jX_{j-1}+U_j,\quad
b_j\sim\operatorname{Unif}\{-0.6,-0.2,0.4,0.8\},\quad
U_j\sim\mathcal N(0,(j+2)/4).
\]
Indices are one-based and noises are independent. The family consists of the $d$ no-collider orientations of the chain, with generating root $X_1$. For each DAG, population regression on its parent sets determines coefficients and residual variances from the same covariance matrix. Thus these population models share the observational distribution.

\paragraph{Nonlinear models and classifier.}
The nonlinear-SCM setting transforms a compatible Gaussian family coordinate-wise, $X_j=\phi(Z_j)$ with $\phi(z)=z+0.1z^3$, so that in graph $\mathcal G$
\[
X_j=\phi\!\left(\sum_{k\in\mathrm{pa}_{\mathcal G}(j)}B_{\mathcal G,jk}\phi^{-1}(X_k)+U_j\right).
\]
An observed intervention at $\theta_j$ sets the latent value to $\phi^{-1}(\theta_j)$. The separate nonlinear-classifier study keeps Gaussian SCMs and uses a linear-plus-$\tanh$ feature map, not an MLP. 

\paragraph{Classifier training and factual instances.}
In the population studies, we generate 2,000 training observations and 10,000 held-out candidates, then choose the first 100 denied factual instances per setup. Training labels satisfy
\[
Y\mid X\sim\operatorname{Bernoulli}\!\left(\sigma\left(2.5\sum_jX_j/r\right)\right),\quad
r=\max\{2000^{-1}\sum_i|\sum_jX_j^i|,10^{-8}\}.
\]
The fitted classifier is logistic regression on training-standardized features, or on those features and their $\tanh$ values. It minimizes mean cross-entropy plus $10^{-3}$ times the squared non-intercept coefficients. It is fixed throughout recourse evaluation.

\paragraph{Actions and costs.}
For the multi-model population experiments, let $R_j$ be the training range. Then
\[
\mathfrak I=\{\{1\},\{d\},\{1,d\}\},\quad
\Theta_{\mathcal I}=\prod_{j\in\mathcal I}[x_j^{\texttt F},x_j^{\texttt F}+\min\{3,R_j/2\}],\quad
\costF(a)=\sum_{j\in\mathcal I}\rho_j|\theta_j-x_j^{\texttt F}|,
\]
with endpoint weights $\rho_1=1$ and $\rho_d=4$. These are interventions on feature levels, not additions to structural equations.

\subsection{Supplied-graph ANM and loan models}

\paragraph{Supplied-graph ANM companion.}
The GP comparison also uses the printed Appendix E ANM of~\citet{karimiAlgorithmicRecourseImperfect2020}:
\[
X_1=U_1,\quad X_2=-1+3\sigma(2X_1)+U_2,\quad X_3=-0.05X_1+0.25X_2^2+U_3,
\]
where $U_1\sim\frac12\mathcal N(-2,1.5)+\frac12\mathcal N(1,1)$,
$U_2\sim\mathcal N(0,0.1)$ and $U_3\sim\mathcal N(0,1)$ are independent. Only one graph is supplied. Each action increases one of the three variables by at most $R_j/2$. Population controls and linear regression runs use cost $|\theta_j-x_j^{\texttt F}|/R_j$; the GP posterior run instead uses the Euclidean cost specified in the main text. 

\paragraph{Loan model.}
\cref{fig:loan-scm} shows the loan structure. The seven features are gender ($G_{\mathrm{sex}}$), age ($A$), education ($E$), loan amount ($L$), duration ($D$), income ($I_{\mathrm{inc}}$) and savings ($S$). Their structural equations are:
\begin{figure}[htbp]
\centering
\begin{lrbox}{\revisiontable}% Native TikZ: inherits the manuscript fonts. Eleven structural edges.
\begin{tikzpicture}[>=Latex,
 v/.style={draw,rounded corners=2pt,align=center,font=\small,minimum width=17mm,minimum height=10mm,inner sep=3pt},
 act/.style={v,fill=black!7},every path/.style={line width=.4pt}]
\node[v] (g) at (0,1.4) {Gender\\$G_{\mathrm{sex}}$};
\node[v] (a) at (0,-1.1) {Age\\$A$};
\node[act] (e) at (3,2.1) {Education\\$E$};
\node[act] (l) at (3,-1.8) {Loan amount\\$L$};
\node[act] (i) at (6,1.4) {Income\\$I_{\mathrm{inc}}$};
\node[v] (d) at (6,-1.8) {Duration\\$D$};
\node[act] (s) at (8.7,1.4) {Savings\\$S$};
\draw[->] (g)--(e);
\draw[->] (g)--(l);
\draw[->] (g)--(d);
\draw[->] (g)--(i);
\draw[->] (a)--(e);
\draw[->] (a)--(l);
\draw[->] (a) to[bend left=12] (d);
\draw[->] (a)--(i);
\draw[->] (e)--(i);
\draw[->] (l)--(d);
\draw[->] (i)--(s);
\end{tikzpicture}\end{lrbox}
\ifdim1.05\wd\revisiontable>\linewidth\resizebox{\linewidth}{!}{\usebox{\revisiontable}}\else\scalebox{1.05}{\usebox{\revisiontable}}\fi
\caption{The generating loan SCM used to simulate observations and evaluate true validity. Shaded nodes are permitted intervention targets; the restricted setting excludes savings. Each action changes only one target. All arrows denote causal dependencies; exogenous noise nodes are omitted. Duration is not actionable but can change after an intervention on loan amount. Graph-specific mechanisms are fitted from observations under the candidate families in Appendix~\ref{app:loan-uncertain}.}
\label{fig:loan-scm}
\end{figure}
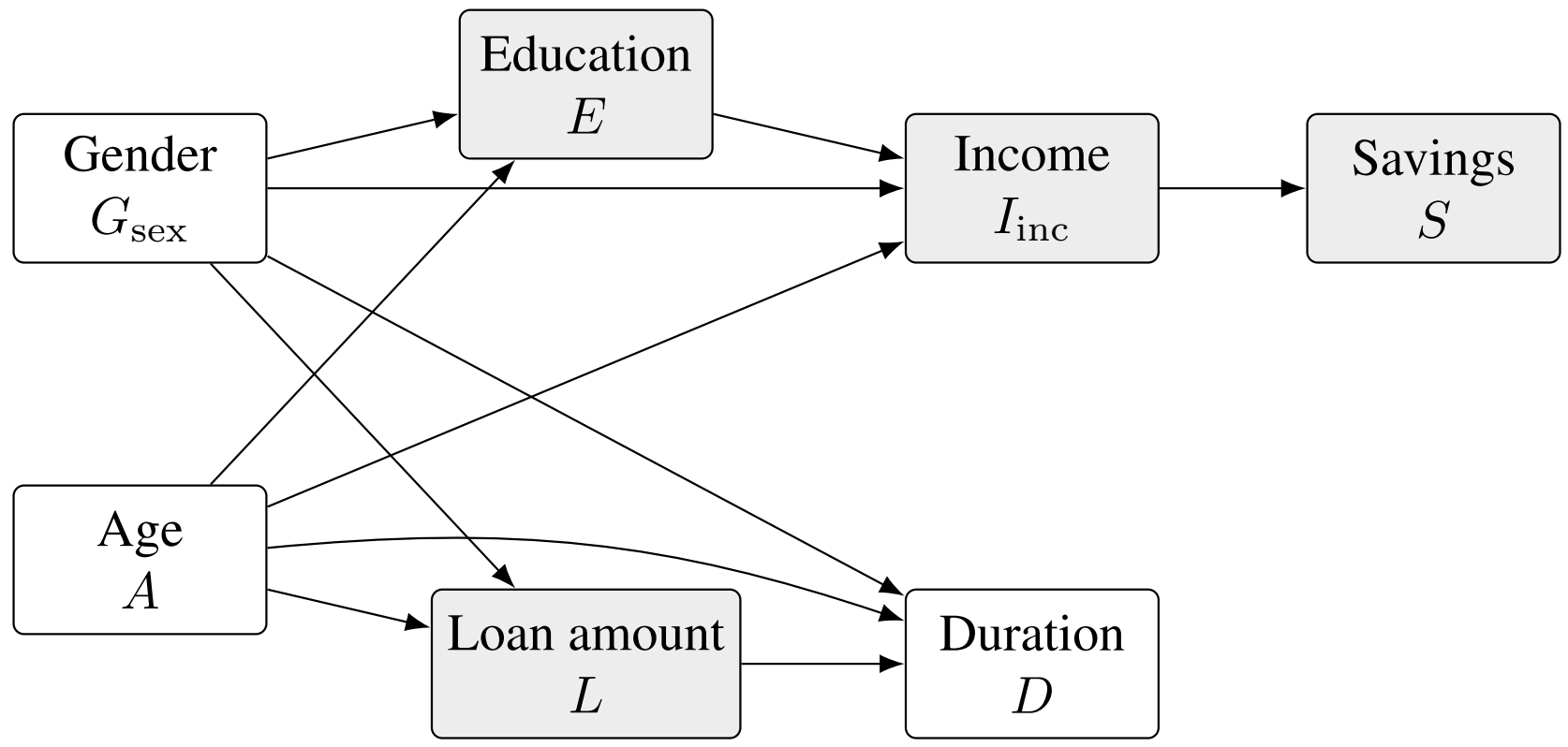

\begin{align}
 G_{\mathrm{sex}}&=U_G,\qquad A=-35+U_A,\\
 E&=-0.5+\sigma(-1+0.5G_{\mathrm{sex}}+\sigma(0.1A)+U_E),\\
 L&=1-0.01(A-5)^2+G_{\mathrm{sex}}+U_L,\\
 D&=-1+0.1A+2G_{\mathrm{sex}}+L+U_D,\\
 I_{\mathrm{inc}}&=-4+0.1(A+35)+2G_{\mathrm{sex}}+G_{\mathrm{sex}}E+U_I,\\
 S&=-4+1.5\max\{I_{\mathrm{inc}},0\}+U_S.
\end{align}
Here $U_G\sim\operatorname{Bernoulli}(0.5)$, $U_A\sim\operatorname{Gamma}(10,3.5)$
(shape, scale), and $(U_E,U_L,U_D,U_I,U_S)$ have independent zero-mean Gaussian
variances $(0.25,4,9,4,25)$. Loan training labels use
\[
Y\mid X\sim\operatorname{Bernoulli}\left(
 \sigma(0.3[-L-D+I_{\mathrm{inc}}+S+I_{\mathrm{inc}}S])\right).
\]
The fitted classifier remains linear logistic regression on standardized features.

\subsection{Replication and numerical references}
Population linear seeds are 3101--3110 for each dimension; nonlinear seeds are 3201--3205 for each of two settings. Each setup contains 100 factual instances. Repeated criterion checks on the same factual instances are not independent samples. Linear references enumerate permitted target sets and solve LP/MILP problems with a conservative $10^{-7}$ margin buffer. Search gaps use the same criterion and action domain. Nonlinear witnesses use 21 levels per action coordinate. Existing coverage intervals bootstrap whole setup seeds 5,000 times.

\subsection{Search implementation}\label{app:implementation}
At the zero-shift initialization, we use the zero subgradient of the Euclidean cost; the implementation also uses zero at the hinge boundary.
With $p$ actionable variables and budget $b$, the number of nonempty target sets is $\sum_{k=1}^{\min(b,p)}\binom{p}{k}$. Our endpoint-intervention studies use three target sets, the supplied-graph ANM uses three singleton targets, and loan uses four singleton targets (three without savings).

\cref{tab:hyperparameters} collects the primary settings. Ablations change only the settings stated in their respective sections. All searches retain the cheapest visited action passing their acceptance check. Deterministic actions are independently replayed; GP actions undergo the separate posterior checks described below. The population configuration sets a 300-second cap per batch/method and a two-second limit per linear reference target problem; capped or unresolved cases remain explicit in reporting.

\begin{table}[H]
\centering\footnotesize
\caption{Recorded primary hyperparameters. OLS denotes fitted linear regression. A dash means not applicable. The smooth target $\tau$ is distinct from \cref{alg:weighted}'s required graph weight $1-\varepsilon$. Updates are per initialization and target set.}
\label{tab:hyperparameters}
\setlength{\tabcolsep}{3pt}\renewcommand{\arraystretch}{1.08}
\begin{tabular*}{\linewidth}{@{\extracolsep{\fill}}lccc@{}}
\toprule
Setting & OLS & GP posterior & Population controls \\
\midrule
Mechanism training observations & 128 & 128 & Supplied mechanisms \\
Step size $\eta$ & 0.001 & 0.001 & 0.001 \\
Penalty $\lambda$ & 200 & 200 & 200 \\
Smooth target $\tau$ & 0.9 & 0.9 & 0.9 \\
Updates $T$ & 500 & 100 & 2,000 \\
Action initializations & \multicolumn{3}{c}{Zero shift, midpoint, and random} \\
Search score & $\sigma(5\operatorname{logit}h)$ & $\sigma(20\hat\Delta_M)$ & $\sigma(5\operatorname{logit}h)$ \\
\cref{alg:worst-case} soft-min $\beta$ & \multicolumn{3}{c}{$\max\{20,2\log|\GG|/0.1\}$ at the primary $\tau=0.9$} \\
\cref{alg:weighted} graph draws $K$ & 32 & 32 & 32 \\
Required graph weight & 0.8 synthetic & 0.8 synthetic; 0.5 loan & 0.8 \\
LCB multiplier $\gamma_{\textsc{lcb}}$ & --- & 2 & --- \\
Search posterior draws $M$ & --- & 16 & --- \\
Verification posterior draws $M'$ & --- & 64 & --- \\
Final posterior evaluation draws & --- & 512 & --- \\
Intervention cost & Weighted $L_1$ & Euclidean & Weighted $L_1$ \\
GP marginal-likelihood starts & --- & 3 & --- \\
\bottomrule
\end{tabular*}
\end{table}

The deterministic surrogate has the same hard approval boundary as $h-0.5$, but changes the search landscape. GP search instead uses the score-LCB margin. Weighted $L_1$ costs permit exact LP references for the linear controls; GP costs remain Euclidean across methods. These are recorded configurations, not claims of optimal hyperparameters. The expanded cohorts retain the original settings without retuning. Loan sweeps use graph weights 0.5, 0.6, 0.8, and 0.9 for both fitted methods. Other graph-weight sweeps and the separate activation, initialization, and smooth-target ablations are reported with their own configurations.

\subsection{Fitting and replication protocol}\label{app:fitted-protocol}
We use fitted linear regression mechanisms as a deterministic baseline and GP posterior mechanisms as our main uncertainty-aware study. Supplied population mechanisms are retained as controls that isolate graph ambiguity. Candidate families are supplied rather than discovered. Both algorithms share the factual instances, classifier, observations, action restrictions, and fitted mechanisms within each setting. The generating SCM is used for evaluation, not as the optimizer's known true model.

For each candidate graph and non-root variable, linear regression fits an intercept and linear coefficients on its parents using 128 training observations. For a factual individual, we infer and retain the fitted residual:
\[
\widehat u_j=x_j^{\texttt F}-\widehat\alpha_{j,\mathcal G}
-\widehat b_{j,\mathcal G}^{\mathsf T}\xF_{\mathrm{pa}_{\mathcal G}(j)}.
\]
After setting $X_{\mathcal I}=\btheta$, we propagate in topological order,
\[
x_j^{\texttt{CF}}=\widehat\alpha_{j,\mathcal G}
+\widehat b_{j,\mathcal G}^{\mathsf T}
\mathbf x^{\texttt{CF}}_{\mathrm{pa}_{\mathcal G}(j)}+\widehat u_j,
\qquad j\notin\mathcal I.
\]
Root variables that are not intervened on retain their factual values. Thus the baseline uses deterministic plug-in counterfactuals. It still has finite-sample estimation error; nonlinear generating mechanisms additionally test misspecification. Linear regression removes posterior sampling from this baseline, not uncertainty about its estimated coefficients.

For linear regression, hard approval is $h(\xCF(a;\mathcal G))\geq0.5$. \cref{tab:hyperparameters} records the different costs and search budgets. These are comparisons of \cref{alg:worst-case,alg:weighted} within each study, not a matched ablation isolating only the effect of posterior uncertainty.

{\leavevmode Every main fitted-model setting has 1,000 denied factual instances: 100 for each of ten seeds. Synthetic seeds are 4101--4110 and loan seeds are 6101--6110. The five original seeds and their first twenty applicants are retained; each cohort is extended to the first 100 denied observations in the same ordered held-out pool. Five new seeds use the same protocol. GP posterior search uses twenty batches of five and linear regression uses five batches of twenty, with fitting shared per seed. Existing batch random streams are unchanged; new batches use separate streams. This extends the earlier cohort rather than creating an entirely independent replication. The synthetic seeded setups can vary the generating SCM, training data, classifier, and evaluation cohort, so across-seed SD reflects variation across whole setups. The five-node setting is evaluated with linear regression. The single-DAG comparison uses these same expanded cohorts. Population controls, exact-reference gaps, and graph-weight and search ablations retain their separately stated cohorts.}

GP mechanisms use an RBF kernel with separate length scales, training-only standardization, and three marginal-likelihood optimization starts. Identical node/parent-set fits are cached across graphs. The procedure conditions on the factual observation and samples the counterfactual posterior. Hyperparameters remain fitted point estimates. In the transformed nonlinear-SCM setting, the generating mechanisms are non-additive. Markov equivalence alone does not ensure that every orientation admits additive-Gaussian GP mechanisms, so fitted results also reflect possible mechanism misspecification.

Posterior search and verification use independent fresh draws at every iterate, with sample counts in \cref{tab:hyperparameters}. Loan additionally tests required graph weights 0.6, 0.8, and 0.9. The separate final evaluation does not change the returned action. The sample standard deviation is a plug-in estimate of posterior spread; it is not an unbiased estimator of the posterior standard deviation.

We report returned coverage, \rev{true validity}, cost, and runtime. \rev{True validity counts a returned action as a true success when} $h(\xCF(a;\mathcal G^\star))\geq0.5$ under the generating SCM. It differs from passing the fitted posterior criterion. Percentages and costs are means and sample standard deviations across fitted-model seeds. Counts are pooled. \rev{True} validity and cost exclude seeds without returns, while coverage includes every seed. Increasing factual instances improves within-fit sampling but does not remove between-fit variation. For the main GP loan comparison, \rev{true} validity and cost use ten seeds with savings. Without savings, Option 1 uses nine seeds for each algorithm; Option 2 uses five for \cref{alg:worst-case} and nine for \cref{alg:weighted}. Coverage uses all ten seeds. Paired costs without savings use nine seeds for Option 1 and five for Option 2.

\subsection{Loan candidate families and replication}
\label{app:loan-uncertain}

We use the generating loan SCM in \cref{fig:loan-scm}, but fit mechanisms from observations
under each supplied candidate graph rather than providing
the true SCM.

\paragraph{Candidate families.}
For Option 1, starting from \cref{fig:loan-scm}, we allow either
direction for education--income and loan amount--duration.
These two covered-edge reversals yield the four-DAG MEC.
All other edges remain fixed. Option 2, the broader eight-DAG family,
additionally allows either direction for income--savings.
Reversing this edge changes collider structure, so the
eight graphs do not form a single MEC. Both families
contain the generating graph and use uniform weights.

In Option 1 (four candidates), required graph weightes
$0.5,0.6,0.8,0.9$ require respectively $2,3,4,4$ candidates
to pass. In Option 2 (eight candidates), they require $4,5,7,8$.
Thus 0.8 and 0.9 match \cref{alg:worst-case}'s hard requirement
in Option 1, although the search objectives
remain different.

\paragraph{Data and fitting.}
For each seed, we generate 2,000 classifier-training
observations and 10,000 held-out observations.
The first 128 training observations fit graph-specific
Linear regression or GP mechanisms using the protocol in
Appendix~\ref{app:fitted-protocol}.
Identical node/parent-set fits are shared across graphs;
reversing an edge changes the fitted parent sets.
A separate 1,000-observation prediction diagnostic set
is not used for tuning.

{\leavevmode Each family/action setting uses 1,000 denied applicants: 100 for each seed 6101--6110.} Algorithms share the
applicants, classifier, fitted mechanisms, and action
restrictions within each setting. Compatible original
batches are retained in this expanded cohort.

\paragraph{Actions and costs.}
Each action increases one of education, loan amount,
income or savings by at most half its training range,
with $\theta_E\leq0.499999$.
The restricted setting removes only savings as a direct
target. Sex, age, and duration remain non-actionable.
Linear regression uses training-range-normalized $L_1$ cost;
GP posterior uses Euclidean cost.

\paragraph{Interpretation.}
Markov equivalence does not guarantee that every candidate
admits equally accurate additive-Gaussian mechanisms.
The generating education mechanism is bounded and
non-additive, and Option 2 includes graphs
outside the generating MEC. These experiments therefore
combine structural uncertainty with mechanism-estimation
error and possible misspecification.
The threshold results and linear regression baseline summary appear in
Appendix~\ref{app:loan-thresholds}.

\section{Additional Results} \label{app:additional}
\subsection{Controlled gradual loss of \rev{coverage under full robustness}}\label{app:graded-motivation}
This controlled experiment uses supplied population mechanisms to isolate structural ambiguity. To match the manuscript's one-based indexing, let
\[
X_1=U_1,\qquad X_{j+1}=\tfrac35X_1+U_{j+1}\quad(j=1,\ldots,19),
\]
where $U_1\sim\mathcal N(0,1)$ and $U_{j+1}\sim\mathcal N(0,1+j/10)$ are independent. The star's MEC contains 20 DAGs, one for each choice of root, with edges oriented away from it. Candidate mechanisms are obtained from the same population covariance.

Fix $h(\mathbf x)=\sigma(w^\top\mathbf x)$ with
\[
w_1=0,\quad(w_2,w_3,w_4,w_5,w_6)=(1/5,2/5,3/5,4/5,1),\quad
w_7=-2,\quad w_8=\cdots=w_{20}=0.
\]
For 100 equally spaced $q\in[1/4,3/2]$, set $x_6^{\texttt F}=-q$ and all other factual coordinates to zero. These are constructed factual instances, not randomly sampled observations. Only $X_1$ is actionable:
$a=do(X_1=\theta)$, $0\leq\theta\leq3$, and $\costF(a)=\theta$.
Because the factual center is zero, $\theta$ also equals its intervention shift.

For a candidate rooted at $r$, the post-action logit is
\[
\ell_r(a)=-q+g_r\theta,\qquad
 g_1=3/5,\quad g_r=\tfrac35(1-w_r)\ (r\ne1).
\]
An intervened center changes every leaf except a non-center root by $3\theta/5$. Since the leaf weights sum to one, the formula follows. Approval is equivalent to $\ell_r(a)\geq0$. Here $\ell_r$ is a deterministic logit, not the GP score-LCB margin $\Delta$.

Order roots as $1,7,8,\ldots,20,2,3,4,5,6$, and let $\GG_L$ contain the first $L$ models. We use $L$ for candidate count to distinguish it from the manuscript's $K$ sampled graphs per update. Only $L=20$ is the complete star MEC. The minimum gains for $L=15,\ldots,20$ are $0.6,0.48,0.36,0.24,0.12,0$, respectively. Thus
\[
\max_{0\leq\theta\leq3}\min_{\mathcal G\in\GG_L}\ell_{\mathcal G}(a)
=3g_{\min,L}-q,
\]
and robust recourse exists exactly when $q\leq3g_{\min,L}$.

With uniform weights renormalized within each prefix, let $g_{[j],L}$ be the $j$th largest gain and $j=\lceil(1-\varepsilon)L\rceil$. Weighted recourse exists exactly when $g_{[j],L}>0$ and $q\leq3g_{[j],L}$; its minimum cost is $q/g_{[j],L}$. These expressions are checked against LP/MILP references.

\begin{table}[H]
\centering\footnotesize
\caption{Returned actions out of 100. Every count agrees
with exact feasibility. All methods return 100 actions
for every prefix $L\leq15$.}
\label{tab:graded-all-methods}
\setlength{\tabcolsep}{3pt}\renewcommand{\arraystretch}{1.08}

\begin{tabular*}{\linewidth}{@{\extracolsep{\fill}}lrrrr@{}}
\toprule
Models $L$ & \cref{alg:worst-case} & Alg. 2: 60\% & 80\% & 90\% \\
\midrule
15 & 100 & 100 & 100 & 100 \\
16 & 95  & 100 & 100 & 100 \\
17 & 66  & 100 & 100 & 95  \\
18 & 38  & 100 & 100 & 66  \\
19 & 9   & 100 & 95  & 38  \\
20 & 0   & 100 & 95  & 38  \\
\bottomrule
\end{tabular*}
\end{table}

We run the existing deterministic search at every $L=1,\ldots,20$, with 2,000 updates, step size $0.001$, penalty 200, sigmoid sharpness 5 applied to the logit, smooth target $\tau=0.9$, 32 sampled graphs per \cref{alg:weighted} update, and zero/midpoint/random initializations. The seed is 20261109; \cref{alg:worst-case} uses $\beta=\max\{20,2\log L/(1-\tau)\}$. This is the deterministic search convention, not the posterior experiment's surrogate or budget. All 7,535 returned actions pass structural replay, and 8,000 LP/MILP checks agree with the analytic feasibility results. No feasible factual instance is missed, and no run hits its cap. Repeated decisions reuse the same 100 factual instances.

The generating center-root model approves the returned actions, while the root-6 model has zero gain and rejects every permitted action. Relaxation therefore sacrifices robustness to a compatible model even when true validity is retained in this construction. The order and classifier are designed; monotonic weighted coverage is not asserted for arbitrary nested families with renormalized weights.

\subsection{Population results: robustness, cost, and coverage}\label{app:population-controls}
We test small linear models, nonlinear classification with linear SCMs, an invertibly transformed nonlinear SCM, and larger linear chains. \Cref{tab:population} compares both algorithms directly. Costs are weighted $L_1$ with $\rho_1=1$ on $X_1$ and $\rho_d=4$ on $X_d$. Every returned \cref{alg:worst-case} action is \rev{a true success under the generating} population SCM. \cref{alg:weighted} increases coverage, but $(1-\varepsilon)$-robustness need not preserve \rev{true validity}.

\begin{table}[H]
\centering\footnotesize
\caption{Population-model results with known mechanisms.
Entries are mean $\pm$ sample standard deviation across
ten runs for each linear setting and five runs for each
nonlinear setting, with 100 factual instances per run.
Within each run, true validity and mean weighted $L_1$
cost are computed among returned actions.
Runs receive equal weight.
\cref{alg:weighted} uses uniform DAG weights and requires
graph weight $1-\varepsilon=0.8$.
Costs are not paired comparisons; standard deviations
describe between-run variability, not confidence intervals.}
\label{tab:population}
\setlength{\tabcolsep}{3pt}\renewcommand{\arraystretch}{1.08}

\begin{tabular*}{\linewidth}{@{\extracolsep{\fill}}lrrrrrr@{}}
\toprule
& \multicolumn{3}{c}{\cref{alg:worst-case}}
& \multicolumn{3}{c}{\cref{alg:weighted}} \\
\cmidrule(lr){2-4}\cmidrule(l){5-7}
Setting
& \shortstack[r]{Coverage\\(\%)}
& \shortstack[r]{True\\validity (\%)}
& Cost
& \shortstack[r]{Coverage\\(\%)}
& \shortstack[r]{True\\validity (\%)}
& Cost \\
\midrule
Linear, 3 nodes
& \ms{91.7}{4.3} & \ms{100.0}{0.0} & \ms{4.91}{0.35}
& \ms{93.6}{4.3} & \ms{84.4}{7.8} & \ms{4.74}{0.29} \\
Nonlinear classifier
& \ms{95.6}{1.8} & \ms{100.0}{0.0} & \ms{4.74}{0.19}
& \ms{97.8}{2.3} & \ms{80.4}{3.7} & \ms{4.52}{0.22} \\
Nonlinear SCM
& \ms{85.4}{4.6} & \ms{100.0}{0.0} & \ms{4.68}{0.27}
& \ms{88.4}{4.0} & \ms{75.3}{5.6} & \ms{4.76}{0.34} \\
\midrule
Linear, 5 nodes
& \ms{83.4}{12.3} & \ms{100.0}{0.0} & \ms{3.93}{1.57}
& \ms{89.0}{5.3} & \ms{45.5}{40.0} & \ms{3.11}{0.88} \\
Linear, 10 nodes
& \ms{56.1}{14.5} & \ms{100.0}{0.0} & \ms{5.68}{1.20}
& \ms{61.8}{12.1} & \ms{36.5}{44.9} & \ms{5.15}{1.33} \\
Linear, 20 nodes
& \ms{36.3}{12.9} & \ms{100.0}{0.0} & \ms{5.42}{1.88}
& \ms{44.7}{6.4} & \ms{32.2}{46.8} & \ms{5.52}{0.84} \\
\bottomrule
\end{tabular*}
\end{table}

% \cref{tab:paired-costs} compares costs only on jointly served applicants. These paired differences answer whether relaxation makes recommendations cheaper for the same people. 

% \begin{table}[H]
% \centering
% \caption{Paired action costs at required robustness 0.8. Negative differences mean Algorithm 2 is cheaper. Each row uses its own common served set.}
% \label{tab:paired-costs}
% \renewcommand{\arraystretch}{1.18}
% \begin{tabular*}{\linewidth}{@{\extracolsep{\fill}}p{0.29\linewidth}rrrr@{}}
% \toprule
% Setting & Pairs & Alg. 1 & Alg. 2 & \shortstack{Alg. 2\\minus 1} \\
% \midrule
% Linear, 3 nodes & 917 & 4.907 & 4.593 & -0.315 \\
% Linear, 5 nodes & 834 & 3.748 & 2.648 & -1.100 \\
% Linear, 10 nodes & 561 & 5.560 & 4.307 & -1.253 \\
% Linear, 20 nodes & 363 & 5.484 & 4.236 & -1.248 \\
% Nonlinear classifier & 478 & 4.742 & 4.349 & -0.393 \\
% Nonlinear SCM & 427 & 4.682 & 4.483 & -0.199 \\
% \bottomrule
% \end{tabular*}
% \end{table}

\subsection{Five passing conditions} \label{app:criteria}
For the deterministic population controls only, define
\[
R_\pi(a)=\sum_{\mathcal G\in\GG}\pi_{\mathcal G}\mathbf1\{h(\xCF(a;\mathcal G))\geq0.5\},\qquad
B_\pi(a)=\sum_{\mathcal G\in\GG}\pi_{\mathcal G}\mathbf1\{h(\xCF(a;\mathcal G))<0.2\}.
\]
\begin{align*}
\mathcal A_1&=\{a\in\AF:\min_{\mathcal G}h(\xCF(a;\mathcal G))\geq0.5\},\\
\mathcal A_2&=\{a\in\AF:R_\pi(a)\geq0.8\},\\
\mathcal A_3&=\{a\in\mathcal A_2:\min_{\mathcal G}h(\xCF(a;\mathcal G))\geq0.2\},\\
\mathcal A_4&=\{a\in\mathcal A_2:B_\pi(a)\leq0.05\},\\
\mathcal A_5&=\{a\in\AF:\sum_{\mathcal G}\pi_{\mathcal G}h(\xCF(a;\mathcal G))\geq0.5\}.
\end{align*}
The main paper uses $\mathcal A_1$ and $\mathcal A_2$. The other conditions test whether a score floor or tail constraint changes the trade-off. $\mathcal A_5$ is a mean-score comparator, not \cref{alg:weighted}'s $(1-\varepsilon)$-robustness objective. $\mathcal A_3$ and $\mathcal A_4$ use separate retention checks along \cref{alg:weighted} trajectories. Costs below are unpaired, since the returned sets can differ.

\definecolor{stdgray}{gray}{0.60}
\providecommand{\ms}[2]{%
  \ensuremath{#1\,{\color{stdgray}\pm #2}}%
}

\newlength{\criteriaNumWidth}
% ============================================================
% Population mechanisms
% ============================================================
\begin{table}[H]
\centering\footnotesize
\caption{Five acceptance criteria under supplied population
mechanisms. Entries are mean $\pm$ sample standard deviation
across ten seeds, with 100 factual instances per seed.
True validity and mean cost are computed among returned actions
within each seed; seeds receive equal weight.
Costs are not paired comparisons.
Gray values indicate standard deviations, not confidence intervals.}
\label{tab:criteria-population}
\setlength{\tabcolsep}{3pt}\renewcommand{\arraystretch}{1.08}

\begin{tabular*}{\linewidth}{@{\extracolsep{\fill}}llrrrrr@{}}
\toprule
Setting & Metric
& $\mathcal A_1$ & $\mathcal A_2$ & $\mathcal A_3$
& $\mathcal A_4$ & $\mathcal A_5$ \\
\midrule

Linear, 3 nodes
& Coverage (\%)
& \ms{91.7}{4.3} & \ms{93.6}{4.3} & \ms{93.6}{4.3}
& \ms{93.6}{4.3} & \ms{99.3}{1.1} \\
& True validity (\%)
& \ms{100.0}{0.0} & \ms{84.4}{7.8} & \ms{84.4}{7.8}
& \ms{84.4}{7.8} & \ms{4.0}{5.8} \\
& Mean cost
& \ms{4.91}{0.35} & \ms{4.74}{0.29} & \ms{4.74}{0.29}
& \ms{4.74}{0.29} & \ms{3.75}{0.93} \\
\midrule

Linear, 5 nodes
& Coverage (\%)
& \ms{83.4}{12.3} & \ms{89.0}{5.3} & \ms{87.2}{8.8}
& \ms{87.2}{8.8} & \ms{94.0}{5.7} \\
& True validity (\%)
& \ms{100.0}{0.0} & \ms{45.5}{40.0} & \ms{47.1}{38.2}
& \ms{47.1}{38.2} & \ms{32.3}{46.9} \\
& Mean cost
& \ms{3.93}{1.57} & \ms{3.11}{0.88} & \ms{3.32}{1.07}
& \ms{3.32}{1.07} & \ms{3.18}{0.80} \\
\midrule

Linear, 10 nodes
& Coverage (\%)
& \ms{56.1}{14.5} & \ms{61.8}{12.1} & \ms{60.7}{12.2}
& \ms{60.7}{12.2} & \ms{65.9}{13.2} \\
& True validity (\%)
& \ms{100.0}{0.0} & \ms{36.5}{44.9} & \ms{36.5}{44.9}
& \ms{36.5}{44.9} & \ms{32.1}{47.2} \\
& Mean cost
& \ms{5.68}{1.20} & \ms{5.15}{1.33} & \ms{5.15}{1.32}
& \ms{5.15}{1.32} & \ms{5.22}{1.18} \\
\midrule

Linear, 20 nodes
& Coverage (\%)
& \ms{36.3}{12.9} & \ms{44.7}{6.4} & \ms{44.7}{6.4}
& \ms{44.7}{6.4} & \ms{44.7}{7.0} \\
& True validity (\%)
& \ms{100.0}{0.0} & \ms{32.2}{46.8} & \ms{32.2}{46.8}
& \ms{32.2}{46.8} & \ms{32.5}{46.8} \\
& Mean cost
& \ms{5.42}{1.88} & \ms{5.52}{0.84} & \ms{5.52}{0.84}
& \ms{5.52}{0.84} & \ms{5.34}{0.97} \\

\bottomrule
\end{tabular*}
\end{table}

\subsection{Activation and initialization ablations} \label{app:search-ablation}
The population activation and initialization comparisons use 150 shared factual instances (five seeds 3401--3405, 30 factual instances each). The softplus alternative is $s(m)=1-\log(1+e^{-5m})$, with derivative $5\sigma(-5m)$; it is unbounded below and is not a probability. Hard acceptance is unchanged. \Cref{tab:search-ablation} compares costs on the factual instances served by both methods. The baseline uses sigmoid search with three starts. Sigmoid remains the primary search activation; softplus is an ablation.

\begin{table}[H]
\centering\footnotesize
\caption{Search ablations. Each algorithm's baseline is
sigmoid search with three starts. Pairs counts factual instances
served by both the variant and its baseline. Paired cost
change is variant minus baseline; positive values mean
higher cost.}
\label{tab:search-ablation}
\setlength{\tabcolsep}{3pt}\renewcommand{\arraystretch}{1.08}

\begin{tabular*}{\linewidth}{@{\extracolsep{\fill}}lrrrrrr@{}}
\toprule
& \multicolumn{3}{c}{\cref{alg:worst-case}}
& \multicolumn{3}{c}{\cref{alg:weighted}} \\
\cmidrule(lr){2-4}\cmidrule(l){5-7}
Variant
& Returned & Pairs & \shortstack[r]{Paired cost\\change}
& Returned & Pairs & \shortstack[r]{Paired cost\\change} \\
\midrule
Sigmoid, three starts
& 122/150 & -- & --
& 130/150 & -- & -- \\
Softplus, three starts
& 124/150 & 122 & +0.160
& 136/150 & 130 & +0.185 \\
One zero start
& 60/150 & 60 & +0.063
& 60/150 & 60 & +0.014 \\
One start, equal updates
& 60/150 & 60 & +0.042
& 60/150 & 60 & +0.014 \\
\bottomrule
\end{tabular*}
\end{table}

\subsection{Graph-weight sensitivity} \label{app:weights}
We evaluate \cref{alg:weighted} on 500 factual instances across five
20-node chains (seeds 3601--3605). Permitted targets are
$X_5,X_{10},X_{15}$, with respective cost weights $(1,2,4)$
and at most two simultaneous targets. Within each setup,
all weight scenarios use the same factual instances, classifier,
action bounds, and search settings.

We compare uniform weights with two scenarios placing
\rev{weight} $0.8$ on one graph and distributing the remaining
\rev{weight} uniformly over the other 19 graphs.
Truth-favoring weights favor the generating graph.
Optimistic incorrect weights favor an incorrect graph
selected to maximize its overstatement of the true
classifier-margin gain over permitted actions with cost
at most one. Selection occurs before factual instance evaluation;
the one-unit budget applies only to graph selection.
Because this construction uses the generating model,
it is a controlled stress test, not an estimated posterior. The graph weights specify importance rather than calibrated probabilities that each graph is true.
The cohort is separate from the main study of 1,000 factual instances.

\begin{table}[H]
\centering\footnotesize
\caption{Graph-weight sensitivity under interior interventions.
True / returned counts actions that succeed under the
generating SCM among returned actions. Missed counts
numerically feasible factual instances for whom search returns
no action.}
\label{tab:weights}
\setlength{\tabcolsep}{3pt}\renewcommand{\arraystretch}{1.08}

\begin{tabular*}{\linewidth}{@{\extracolsep{\fill}}lrrrr@{}}
\toprule
Weights & $1-\varepsilon$ & Returned
& \shortstack[r]{True /\\returned}
& Missed \\
\midrule
Uniform              & 0.6 & 215/500 & 123/215 & 0  \\
Uniform              & 0.8 & 205/500 & 162/205 & 0  \\
Uniform              & 0.9 & 191/500 & 191/191 & 0  \\
\midrule
Truth favoring       & 0.6 & 236/500 & 236/236 & 12 \\
Truth favoring       & 0.8 & 236/500 & 236/236 & 12 \\
Truth favoring       & 0.9 & 191/500 & 191/191 & 5  \\
\midrule
Optimistic incorrect & 0.6 & 258/500 & 38/258  & 13 \\
Optimistic incorrect & 0.8 & 258/500 & 38/258  & 13 \\
Optimistic incorrect & 0.9 & 212/500 & 115/212 & 3  \\
\bottomrule
\end{tabular*}
\end{table}

At required \rev{graph weight} $0.8$, optimistic incorrect weighting
increases returned actions from 205 to 258, but reduces
true successes from 162 to 38. Thus, greater coverage
under the specified weights need not imply greater
success under the generating model. This deliberately
constructed example demonstrates a risk of misleading
weights, rather than a claim that incorrect weights
always worsen performance.

\begingroup
\subsection{Pooled fitted-study counts}\label{app:fitted-details}
Counts pool factual instances across fitted seeds; they are not equal-weight seed-mean percentages. { Each linear regression synthetic row and each GP posterior row has 1,000 factual instances.} New-sample validity counts and costs already appear in the main tables and are not duplicated here.
\begin{table}[H]
\centering\footnotesize
\caption{Pooled returned and true-success counts. True success conditions on returned actions. \cref{alg:weighted} requires \rev{graph weight} 0.8.}
\label{tab:fitted-counts}
\setlength{\tabcolsep}{3pt}\renewcommand{\arraystretch}{1.08}

\begin{tabular*}{\linewidth}{@{\extracolsep{\fill}}lrrrr@{}}\toprule
& \multicolumn{2}{c}{\cref{alg:worst-case}} & \multicolumn{2}{c}{\cref{alg:weighted}} \\
\cmidrule(lr){2-3}\cmidrule(l){4-5}
Setting & Returned & \shortstack[r]{True/\\returned} & Returned & \shortstack[r]{True/\\returned} \\ \midrule
Linear regression: Linear, 3 & { 920/1000} & { 824/920} & { 942/1000} & { 676/942} \\
Linear regression: Nonlinear classifier & { 921/1000} & { 835/921} & { 957/1000} & { 677/957} \\
Linear regression: Nonlinear SCM & { 811/1000} & { 743/811} & { 849/1000} & { 627/849} \\
Linear regression: Supplied-graph ANM & { 995/1000} & { 645/995} & { 995/1000} & { 645/995} \\
Linear regression: Linear, 5 & { 823/1000} & { 695/823} & { 902/1000} & { 416/902} \\
Linear regression: Linear, 10 & { 612/1000} & { 495/612} & { 697/1000} & { 294/697} \\
Linear regression: Linear, 20 & { 358/1000} & { 274/358} & { 439/1000} & { 171/439} \\
GP: Linear, 3 & { 906/1000} & { 896/906} & { 925/1000} & { 882/925} \\
GP: Nonlinear classifier & { 903/1000} & { 897/903} & { 933/1000} & { 897/933} \\
GP: Nonlinear SCM & { 792/1000} & { 774/792} & { 822/1000} & { 770/822} \\
GP: Supplied-graph ANM & { 893/1000} & { 888/893} & { 893/1000} & { 888/893} \\
GP: Linear, 10 & { 548/1000} & { 507/548} & { 688/1000} & { 349/688} \\
GP: Linear, 20 & { 341/1000} & { 251/341} & { 436/1000} & { 184/436} \\
\bottomrule
\end{tabular*}
\end{table}\FloatBarrier
\subsection{Loan thresholds and linear regression baseline}\label{app:loan-thresholds}
\cref{tab:loan-sweep} reports the GP posterior sweep without repeating the main cost and seed-mean tables. Four equal graph weights require 2, 3, 4, and 4 passes at \rev{required graph weights} 0.5, 0.6, 0.8, and 0.9; eight require 4, 5, 7, and 8. Equal hard conditions need not produce identical actions or new-sample validity outcomes under stochastic search.
\begin{table}[H]
\centering\footnotesize
\caption{GP loan thresholds. Each family/action setting has 1,000 factual instances across ten seeds. \cref{alg:worst-case} is the same reference within each family/action block; its counts are repeated for comparison. New-sample validity (passed/returned) uses independent evaluation on new samples from the fitted causal models. New-sample validity percentages are computed from pooled passed/returned counts, not averaged across seeds.}
\label{tab:loan-sweep}
\setlength{\tabcolsep}{3pt}\renewcommand{\arraystretch}{1.08}

\begin{tabular*}{\linewidth}{@{\extracolsep{\fill}}lrrrrrr@{}}\toprule
& \multicolumn{3}{c}{\cref{alg:worst-case}} & \multicolumn{3}{c}{\cref{alg:weighted}} \\
\cmidrule(lr){2-4}\cmidrule(l){5-7}
$1-\varepsilon$ & Returned & True/ret. & \shortstack[r]{New-sample validity\\passed/returned (\%)} & Returned & True/ret. & \shortstack[r]{New-sample validity\\passed/returned (\%)} \\ \midrule
\multicolumn{7}{@{}l}{\textit{Option 1 (four-DAG MEC); direct savings allowed}} \\
0.5 & { 937/1000} & { 925/937} & { 912/937} & { 937/1000} & { 909/937} & { 890/937} \\
0.6 & { 937/1000} & { 925/937} & { 912/937} & { 937/1000} & { 921/937} & { 901/937} \\
0.8 & { 937/1000} & { 925/937} & { 912/937} & { 937/1000} & { 925/937} & { 913/937} \\
0.9 & { 937/1000} & { 925/937} & { 912/937} & { 937/1000} & { 925/937} & { 913/937} \\
\midrule
\multicolumn{7}{@{}l}{\textit{Option 1 (four-DAG MEC); direct savings excluded}} \\
0.5 & { 200/1000} & { 186/200} & { 149/200} & { 225/1000} & { 191/225} & { 129/225} \\
0.6 & { 200/1000} & { 186/200} & { 149/200} & { 214/1000} & { 191/214} & { 130/214} \\
0.8 & { 200/1000} & { 186/200} & { 149/200} & { 204/1000} & { 189/204} & { 147/204} \\
0.9 & { 200/1000} & { 186/200} & { 149/200} & { 204/1000} & { 189/204} & { 147/204} \\
\midrule
\multicolumn{7}{@{}l}{\textit{Option 2 (eight-DAG broader family); direct savings allowed}} \\
0.5 & { 922/1000} & { 921/922} & { 900/922} & { 933/1000} & { 922/933} & { 919/933} \\
0.6 & { 922/1000} & { 921/922} & { 900/922} & { 929/1000} & { 928/929} & { 782/929} \\
0.8 & { 922/1000} & { 921/922} & { 900/922} & { 928/1000} & { 927/928} & { 863/928} \\
0.9 & { 922/1000} & { 921/922} & { 900/922} & { 928/1000} & { 927/928} & { 902/928} \\
\midrule
\multicolumn{7}{@{}l}{\textit{Option 2 (eight-DAG broader family); direct savings excluded}} \\
0.5 & { 19/1000} & { 17/19} & { 15/19} & { 202/1000} & { 186/202} & { 152/202} \\
0.6 & { 19/1000} & { 17/19} & { 15/19} & { 31/1000} & { 29/31} & { 22/31} \\
0.8 & { 19/1000} & { 17/19} & { 15/19} & { 21/1000} & { 19/21} & { 15/21} \\
0.9 & { 19/1000} & { 17/19} & { 15/19} & { 19/1000} & { 17/19} & { 15/19} \\
\bottomrule
\end{tabular*}
\end{table}\FloatBarrier
{\leavevmode For the linear regression loan baseline under Option 1, allowing direct savings, \cref{alg:worst-case} returns 970/1000 actions with 970 true successes. \cref{alg:weighted} returns 970, 970, 970, 970 actions with 970, 970, 970, 970 true successes at \rev{required graph weights} 0.5, 0.6, 0.8, and 0.9, respectively. For the linear regression loan baseline under Option 1, excluding direct savings, \cref{alg:worst-case} returns 522/1000 actions with 392 true successes. \cref{alg:weighted} returns 525, 522, 522, 522 actions with 393, 392, 392, 392 true successes at \rev{required graph weights} 0.5, 0.6, 0.8, and 0.9, respectively. For the linear regression loan baseline under Option 2, allowing direct savings, \cref{alg:worst-case} returns 967/1000 actions with 967 true successes. \cref{alg:weighted} returns 968, 967, 967, 967 actions with 968, 967, 967, 967 true successes at \rev{required graph weights} 0.5, 0.6, 0.8, and 0.9, respectively. For the linear regression loan baseline under Option 2, excluding direct savings, \cref{alg:worst-case} returns 25/1000 actions with 25 true successes. \cref{alg:weighted} returns 505, 36, 25, 25 actions with 379, 33, 25, 25 true successes at \rev{required graph weights} 0.5, 0.6, 0.8, and 0.9, respectively. Linear regression uses range-normalized $L_1$ cost, so its numerical costs are not compared with GP Euclidean costs.}

\subsection{Paired fitted-study costs}\label{app:fitted-paired}
For each seed, we compute each algorithm's mean cost and their mean difference on the same jointly served factual instances. We then report the mean and sample SD across contributing seeds; seeds without jointly served factual instances are omitted. Negative values favor \cref{alg:weighted}. These are paired differences, not optimality gaps. Each panel uses its own cost definition.
\begin{table}[H]
\centering\footnotesize
\caption{Paired intervention costs: required \rev{graph weight} 0.8 for synthetic settings and 0.5 for loan. Costs use only factual instances served by both algorithms within each seed. Entries are equal-weight seed means $\pm$ sample SD; jointly served counts are pooled. Differences are \cref{alg:weighted} minus \cref{alg:worst-case}; negative values indicate lower cost for \cref{alg:weighted}. Paired summaries use ten seeds except the GP loan rows without savings: nine for Option 1 and five for Option 2.}
\label{tab:fitted-pairs}
\setlength{\tabcolsep}{3pt}\renewcommand{\arraystretch}{1.08}

\begin{tabular*}{\linewidth}{@{\extracolsep{\fill}}lrrrr@{}}
\toprule
Setting & \shortstack[r]{Jointly\\served} & \shortstack[r]{\cref{alg:worst-case}\\paired cost} & \shortstack[r]{\cref{alg:weighted}\\paired cost} & \shortstack[r]{Difference\\A2 minus A1} \\
\midrule
Linear regression: Linear, 3 nodes & { 920} & { \ms{4.96}{0.40}} & { \ms{4.62}{0.40}} & { \ms{-0.34}{0.10}} \\
Linear regression: Nonlinear classifier & { 921} & { \ms{4.99}{0.48}} & { \ms{4.64}{0.48}} & { \ms{-0.35}{0.08}} \\
Linear regression: Nonlinear SCM & { 811} & { \ms{5.15}{0.45}} & { \ms{4.88}{0.46}} & { \ms{-0.27}{0.16}} \\
Linear regression: Supplied-graph ANM & { 995} & { \ms{0.19}{0.01}} & { \ms{0.19}{0.01}} & { \ms{0.00}{0.00}} \\
Linear regression: Linear, 5 nodes & { 823} & { \ms{4.77}{1.84}} & { \ms{2.85}{0.92}} & { \ms{-1.92}{2.26}} \\
Linear regression: Linear, 10 nodes & { 612} & { \ms{5.37}{1.39}} & { \ms{3.43}{1.41}} & { \ms{-1.94}{2.13}} \\
Linear regression: Linear, 20 nodes & { 358} & { \ms{6.27}{1.25}} & { \ms{4.43}{1.43}} & { \ms{-1.84}{1.73}} \\
GP: Linear, 3 nodes & { 906} & { \ms{1.33}{0.10}} & { \ms{1.28}{0.08}} & { \ms{-0.05}{0.05}} \\
GP: Nonlinear classifier & { 903} & { \ms{1.34}{0.10}} & { \ms{1.28}{0.08}} & { \ms{-0.06}{0.06}} \\
GP: Nonlinear SCM & { 792} & { \ms{1.38}{0.14}} & { \ms{1.31}{0.11}} & { \ms{-0.07}{0.06}} \\
GP: Supplied-graph ANM & { 893} & { \ms{1.99}{0.17}} & { \ms{1.99}{0.17}} & { \ms{0.00}{0.00}} \\
GP: Linear, 10 nodes & { 548} & { \ms{2.01}{0.45}} & { \ms{1.40}{0.21}} & { \ms{-0.61}{0.57}} \\
GP: Linear, 20 nodes & { 341} & { \ms{2.23}{0.22}} & { \ms{1.72}{0.39}} & { \ms{-0.51}{0.28}} \\
GP: Loan Option 1 / savings & { 937} & { \ms{6.60}{0.56}} & { \ms{6.53}{0.60}} & { \ms{-0.07}{0.08}} \\
GP: Loan Option 1 / no savings & { 200} & { \ms{3.10}{1.24}} & { \ms{2.88}{1.29}} & { \ms{-0.22}{0.11}} \\
GP: Loan Option 2 / savings & { 922} & { \ms{7.06}{0.59}} & { \ms{6.57}{0.56}} & { \ms{-0.49}{0.34}} \\
GP: Loan Option 2 / no savings & { 19} & { \ms{0.29}{0.08}} & { \ms{0.20}{0.06}} & { \ms{-0.09}{0.10}} \\
\bottomrule
\end{tabular*}
\end{table}\FloatBarrier
\begingroup

\subsection{Matched single-DAG recourse}\label{app:single-dag}
For each factual instance, we draw one DAG uniformly from the supplied candidate family, independently of the generating graph and outcomes, and run the same search on that DAG alone. All methods use the same ten seeds and 100 factual instances per seed, training observations, fitted mechanisms, classifier, action bounds, costs, and search settings. The baseline has independent initialization and Monte Carlo streams; it is not the generating-SCM oracle. There is one prescribed graph draw per factual instance, with no outcome-based graph selection. We retain the original five seeds' first twenty instances and their baseline outputs, extending to the same 1,000-instance cohorts as the main tables. Across 21 settings, this gives 21,000 baseline evaluations: 2,100 reused and 18,900 new. All 15,574 returned baseline actions passed the saved replay checks against their selected-graph criterion.

We evaluate every returned action under the generating SCM and across the full candidate family. The latter check uses deterministic fitted margins for linear regression and an independent 512-draw posterior evaluation for GP. The posterior evaluation does not change the chosen action and is not a statistical certificate. Full-family passes require every graph to pass, including for \cref{alg:weighted}; they differ from its weighted new-sample validity in the main tables. \cref{tab:single-ols,tab:single-posterior} report pooled counts, with each outcome's denominator explicit. In the one-graph ANM control, the methods have the same hard requirement; differences arise from search and Monte Carlo streams, not graph ambiguity.

On 20-node GP chains, single-DAG recourse returns 433 actions, with 181 true successes and 55 fresh full-family passes. \cref{alg:worst-case} returns 341, with 251 true successes and 229 full-family passes; \cref{alg:weighted} returns 436, with 184 true successes and 53 full-family passes. Thus \cref{alg:worst-case} improves \rev{full-family validity} and true \rev{successes} here while serving fewer applicants. \cref{alg:weighted} performs similarly to single-DAG recourse on this setting, rather than uniformly outperforming it. In the eight-graph loan setting without direct savings, the returned/true-success counts are 143/121 for single-DAG search, 19/17 for \cref{alg:worst-case}, and 202/186 for \cref{alg:weighted}. The corresponding full-family pass counts are 14, 15, and 13: \cref{alg:weighted}'s gain in coverage and true successes does not imply full-family \rev{validity}.

These comparisons support the \rev{robustness}--coverage--cost trade-off. In the three-node linear GP setting, \cref{alg:worst-case} achieves 896 true successes and \cref{alg:weighted} achieves 882, versus 557 for single-DAG recourse, despite returning fewer actions. Their paired costs exceed the baseline's by $\ms{0.14}{0.05}$ and $\ms{0.10}{0.03}$, respectively. On 20-node GP chains, \cref{alg:worst-case}'s paired cost difference is $\ms{0.50}{0.27}$, whereas \cref{alg:weighted}'s is $\ms{-0.01}{0.02}$. Neither algorithm dominates the baseline in every setting or metric.

Recorded timings retain selected-graph search, verification, final evaluation, full-family evaluation, and replay separately. Shared fitting/setup is reused from the primary experiment and charged once per seed/setting when forming recorded totals. Legacy batch timings are retained unchanged. Batch sizes and execution periods differ from the primary runs, so these timings are not a controlled speed comparison.

\begin{table}[t]
\centering\footnotesize
\caption{Matched single-DAG comparison: fitted linear regression. Each setting uses ten seeds and 100 factual instances per seed. Cells give returned / true-success / full-family-pass counts, all out of 1,000 evaluated instances. True successes and full-family passes are subsets of returns. Full-family passes use deterministic fitted-model margins. \cref{alg:weighted} requires graph weight 0.8 in synthetic settings and 0.5 in loan settings.}
\label{tab:single-ols}
\setlength{\tabcolsep}{3pt}\renewcommand{\arraystretch}{1.08}
\begin{tabular*}{\linewidth}{@{\extracolsep{\fill}}lrrr@{}}\toprule
Setting & Single DAG & \cref{alg:worst-case} & \cref{alg:weighted}\\\midrule
{Linear, 3 nodes} & {963/282/314} & {920/824/920} & {942/676/735}\\
{Nonlinear classifier, 3 nodes} & {959/279/313} & {921/835/921} & {957/677/735}\\
{Nonlinear SCM, 3 nodes} & {881/251/239} & {811/743/811} & {849/627/594}\\
{Supplied-graph ANM} & {991/626/991} & {995/645/995} & {995/645/995}\\
{Linear, 5 nodes} & {879/443/384} & {823/695/823} & {902/416/419}\\
{Linear, 10 nodes} & {679/297/289} & {612/495/612} & {697/294/300}\\
{Linear, 20 nodes} & {438/177/118} & {358/274/358} & {439/171/115}\\
\midrule
{Loan, four graphs + savings} & {973/973/973} & {970/970/970} & {970/970/970}\\
{Loan, four graphs, no savings} & {522/392/352} & {522/392/522} & {525/393/182}\\
{Loan, eight graphs + savings} & {968/968/627} & {967/967/967} & {968/968/340}\\
{Loan, eight graphs, no savings} & {285/225/10} & {25/25/25} & {505/379/0}\\
\bottomrule\end{tabular*}
\end{table}
\begin{table}[t]
\centering\footnotesize
\caption{Matched single-DAG comparison: GP posterior. Each setting uses ten seeds and 100 factual instances per seed. Cells give returned / true-success / full-family-pass counts, all out of 1,000 evaluated instances. True successes and full-family passes are subsets of returns. Full-family passes use an independent 512-draw posterior evaluation. \cref{alg:weighted} requires graph weight 0.8 in synthetic settings and 0.5 in loan settings.}
\label{tab:single-posterior}
\setlength{\tabcolsep}{3pt}\renewcommand{\arraystretch}{1.08}
\begin{tabular*}{\linewidth}{@{\extracolsep{\fill}}lrrr@{}}\toprule
Setting & Single DAG & \cref{alg:worst-case} & \cref{alg:weighted}\\\midrule
{Linear, 3 nodes} & {945/557/433} & {906/896/883} & {925/882/681}\\
{Nonlinear classifier, 3 nodes} & {947/561/437} & {903/897/879} & {933/897/674}\\
{Nonlinear SCM, 3 nodes} & {865/451/283} & {792/774/758} & {822/770/493}\\
{Supplied-graph ANM} & {910/909/890} & {893/888/866} & {893/888/866}\\
{Linear, 10 nodes} & {683/367/171} & {548/507/433} & {688/349/169}\\
{Linear, 20 nodes} & {433/181/55} & {341/251/229} & {436/184/53}\\
\midrule
{Loan, four graphs + savings} & {946/912/873} & {937/925/912} & {937/909/864}\\
{Loan, four graphs, no savings} & {240/199/93} & {200/186/149} & {225/191/90}\\
{Loan, eight graphs + savings} & {924/907/612} & {922/921/900} & {933/922/442}\\
{Loan, eight graphs, no savings} & {143/121/14} & {19/17/15} & {202/186/13}\\
\bottomrule\end{tabular*}
\end{table}

Paired intervention costs on jointly served instances appear in \cref{tab:single-cost}. Means and sample SDs use the contributing seed-level paired differences; seeds without common returns are omitted. These empirical differences include search variation and are not exact-optimum ambiguity premiums.
\begin{table}[t]
\centering\footnotesize
\caption{Paired cost differences against single-DAG search. Entries give primary minus single-DAG cost, as mean $\pm$ sample SD across seeds with common returns; $n$ is the pooled number of jointly served instances. Negative values favor the primary algorithm. OLS uses weighted $L_1$ cost; GP uses Euclidean cost. Costs are not compared across these panels.}
\label{tab:single-cost}
\setlength{\tabcolsep}{3pt}\renewcommand{\arraystretch}{1.08}
\begin{tabular*}{\linewidth}{@{\extracolsep{\fill}}llrrrr@{}}\toprule
Panel & Setting & A1 $-$ single & $n$ & A2 $-$ single & $n$\\\midrule
{OLS} & {Supplied-graph ANM} & {\ms{0.00}{0.00}} & {989} & {\ms{0.00}{0.00}} & {989}\\
{OLS} & {Linear, 10 nodes} & {\ms{1.88}{1.73}} & {602} & {\ms{0.06}{0.37}} & {664}\\
{OLS} & {Linear, 20 nodes} & {\ms{1.81}{1.64}} & {358} & {\ms{0.02}{0.21}} & {434}\\
{OLS} & {Linear, 3 nodes} & {\ms{2.05}{0.38}} & {915} & {\ms{1.74}{0.43}} & {930}\\
{OLS} & {Linear, 5 nodes} & {\ms{1.83}{1.77}} & {809} & {\ms{0.06}{0.44}} & {863}\\
{OLS} & {8 graphs, no savings} & {\ms{0.17}{0.14}} & {25} & {\ms{0.00}{0.04}} & {276}\\
{OLS} & {8 graphs + savings} & {\ms{0.00}{0.00}} & {960} & {\ms{0.00}{0.00}} & {961}\\
{OLS} & {4 graphs, no savings} & {\ms{0.00}{0.00}} & {522} & {\ms{0.00}{0.00}} & {522}\\
{OLS} & {4 graphs + savings} & {\ms{0.00}{0.00}} & {963} & {\ms{0.00}{0.00}} & {963}\\
{OLS} & {Nonlinear classifier} & {\ms{1.90}{0.29}} & {912} & {\ms{1.61}{0.34}} & {941}\\
{OLS} & {Nonlinear SCM} & {\ms{2.09}{0.31}} & {804} & {\ms{1.89}{0.38}} & {829}\\
{GP} & {Supplied-graph ANM} & {\ms{-0.01}{0.03}} & {868} & {\ms{-0.01}{0.03}} & {868}\\
{GP} & {Linear, 10 nodes} & {\ms{0.58}{0.54}} & {546} & {\ms{-0.03}{0.05}} & {655}\\
{GP} & {Linear, 20 nodes} & {\ms{0.50}{0.27}} & {339} & {\ms{-0.01}{0.02}} & {428}\\
{GP} & {Linear, 3 nodes} & {\ms{0.14}{0.05}} & {892} & {\ms{0.10}{0.03}} & {905}\\
{GP} & {8 graphs, no savings} & {\ms{0.07}{0.08}} & {19} & {\ms{-0.01}{0.61}} & {120}\\
{GP} & {8 graphs + savings} & {\ms{0.36}{0.43}} & {903} & {\ms{-0.15}{0.24}} & {911}\\
{GP} & {4 graphs, no savings} & {\ms{0.17}{0.16}} & {199} & {\ms{-0.08}{0.29}} & {219}\\
{GP} & {4 graphs + savings} & {\ms{0.13}{0.25}} & {927} & {\ms{0.06}{0.20}} & {927}\\
{GP} & {Nonlinear classifier} & {\ms{0.15}{0.05}} & {891} & {\ms{0.11}{0.02}} & {913}\\
{GP} & {Nonlinear SCM} & {\ms{0.20}{0.07}} & {785} & {\ms{0.15}{0.03}} & {805}\\
\bottomrule\end{tabular*}
\end{table}

\subsection{Optimization gaps on population controls}\label{app:optimization-gaps}
We reuse the saved linear population-model controls, with supplied mechanisms, weighted $L_1$ costs, and 2,000 search updates. For each returned action $\widehat a$, the absolute gap is $\costF(\widehat a)-c^\star$ and the relative gap is $(\costF(\widehat a)-c^\star)/c^\star$. \cref{alg:worst-case} is compared with the robust LP optimum; \cref{alg:weighted} with the weighted-success LP/MILP optimum at graph weight 0.8, using the same permitted targets and bounds. Relative gaps omit reference costs at or below $10^{-8}$. These gaps condition on returned actions, so we also report numerical infeasibility and feasible cases missed by search. They are distinct from paired cost differences between methods and do not establish optimality for the fitted GP experiments.
\begin{table}[t]

\centering\small
\caption{Saved exact-reference diagnostics on linear population controls. Each row has 1,000 factual instances. Infeasibility is the numerical reference-solver status. Median relative gaps are percentages; mean absolute gaps use the study's weighted $L_1$ units.}
\label{tab:optimization-gaps}

\begin{tabular}{rrrrrrr}\toprule
Algorithm & Nodes & Evaluated & Infeasible & Feasible missed & Median gap (\%) & Mean abs. gap\\\midrule
1 & 3 & 1000 & 50 & 33 & 19.7 & 0.722\\
1 & 5 & 1000 & 124 & 42 & 12.9 & 0.619\\
1 & 10 & 1000 & 421 & 18 & 6.4 & 0.849\\
1 & 20 & 1000 & 637 & 0 & 3.6 & 0.637\\
2 & 3 & 1000 & 33 & 31 & 21.5 & 0.772\\
2 & 5 & 1000 & 77 & 33 & 10.2 & 0.573\\
2 & 10 & 1000 & 347 & 35 & 6.7 & 0.883\\
2 & 20 & 1000 & 553 & 0 & 3.9 & 0.773\\
\bottomrule\end{tabular}
\end{table}

\endgroup

\subsection{Recorded runtime}\label{app:fitted-runtime}

\paragraph{Runtime environment.} Experiments ran locally on an Apple M5 Pro computer with 24\,GiB of memory and 15 CPU cores. The expanded-run environment record reports macOS 26.6.2 (ARM64), Python 3.12.13, NumPy 2.5.4, SciPy 1.18.1, and PyTorch 2.14.1, with one runner thread. These versions describe the recorded expanded-run environment; reused historical timing components retain their original measurements. The timings below are local wall-clock measurements, not an isolated hardware benchmark.

\cref{fig:runtime} compares \cref{alg:worst-case,alg:weighted} within each study, using its timing scope and search budget. Linear regression timings cover fitting and search; GP posterior timings also include verification and final evaluation.

{\leavevmode Each total covers 100 factual instances per fitted seed. Posterior time sums twenty five-instance action batches and counts shared setup/fitting once. It includes per-iterate verification, final evaluation, and measured first result writes, but excludes imports, audit replay, report generation, and cache-copy overhead. Compatible historical timing components are reused as measured. Linear regression records fitting plus the five twenty-instance searches. Four-graph loan timings conservatively charge the shared eight-graph fitting cost. Amortized time divides the total by 100; it is not isolated single-request latency.}
\begin{table}[H]
\centering\footnotesize
\caption{Recorded runtime in seconds: seed mean $\pm$ sample SD. \cref{alg:weighted} uses \rev{required graph weight} 0.8 for linear settings and 0.5 for loan. Linear regression uses ten seeds and fitting-plus-search timing; posterior uses ten seeds and recorded pipeline timing. Amortized values are not single-request measurements.}
\label{tab:runtime-main}
\setlength{\tabcolsep}{3pt}\renewcommand{\arraystretch}{1.08}

\begin{tabular*}{\linewidth}{@{\extracolsep{\fill}}lrrrr@{}}
\toprule
& \multicolumn{2}{c}{\cref{alg:worst-case}} & \multicolumn{2}{c}{\cref{alg:weighted}} \\
\cmidrule(lr){2-3}\cmidrule(l){4-5}
Setting & \shortstack[r]{100-instance\\total (s)} & \shortstack[r]{Amortized\\per instance (s)} & \shortstack[r]{100-instance\\total (s)} & \shortstack[r]{Amortized\\per instance (s)} \\
\midrule
Linear regression, 3 nodes & { \ms{2.96}{0.09}} & { \ms{0.03}{0.00}} & { \ms{2.80}{0.14}} & { \ms{0.03}{0.00}} \\
Linear regression, 10 nodes & { \ms{4.32}{0.35}} & { \ms{0.04}{0.00}} & { \ms{4.24}{0.32}} & { \ms{0.04}{0.00}} \\
Linear regression, 20 nodes & { \ms{7.70}{0.42}} & { \ms{0.08}{0.00}} & { \ms{7.86}{0.41}} & { \ms{0.08}{0.00}} \\
GP posterior, 3 nodes & { \ms{117.04}{1.09}} & { \ms{1.17}{0.01}} & { \ms{116.52}{1.23}} & { \ms{1.17}{0.01}} \\
GP posterior, 10 nodes & { \ms{206.37}{32.86}} & { \ms{2.06}{0.33}} & { \ms{205.28}{33.34}} & { \ms{2.05}{0.33}} \\
GP posterior, 20 nodes & { \ms{403.11}{20.29}} & { \ms{4.03}{0.20}} & { \ms{403.20}{19.82}} & { \ms{4.03}{0.20}} \\
Loan GP, Option 1 / savings & { \ms{67.84}{3.63}} & { \ms{0.68}{0.04}} & { \ms{67.30}{3.61}} & { \ms{0.67}{0.04}} \\
Loan GP, Option 1 / no savings & { \ms{64.62}{3.54}} & { \ms{0.65}{0.04}} & { \ms{63.99}{3.42}} & { \ms{0.64}{0.03}} \\
Loan GP, Option 2 / savings & { \ms{142.65}{7.67}} & { \ms{1.43}{0.08}} & { \ms{142.04}{7.57}} & { \ms{1.42}{0.08}} \\
Loan GP, Option 2 / no savings & { \ms{104.61}{5.93}} & { \ms{1.05}{0.06}} & { \ms{104.49}{6.17}} & { \ms{1.04}{0.06}} \\
\bottomrule\end{tabular*}

\end{table}\FloatBarrier

\subsection{Posterior verification and search-target sensitivity}\label{app:gp-posterior}
The posterior margin is estimated as
\[
\widehat\Delta_M=\bar h_M-2\widehat s_M-0.5,\qquad
\widehat s_M^2=\frac1{M-1}\sum_{m=1}^{M}(h^{(m)}-\bar h_M)^2.
\]
The penalty uses the unscaled score margin. The standard deviation is not a standard error, and the expression is not a calibrated confidence bound. Search and verification use independent fresh streams each iterate; repeated graph indices reuse the same per-graph estimate within an update. The independent 512-draw final check is reported in the new-sample validity (passed/returned) columns and does not modify the chosen action. Repeated noisy checks followed by cheapest-action selection can favor Monte Carlo fluctuations; a failure in the new-sample validity evaluation is not itself proof that the exact posterior criterion fails.

\paragraph{Search-target ablation.}
We vary only the smooth search target $\tau\in\{0.5,0.7,0.8,0.9,0.93\}$. Required graph weight remains $1-\varepsilon=0.8$, and $\beta$ is fixed. These targets are not different hard approval thresholds. This retained retrospective ablation uses the original three fitted seeds and five factual instances per seed in each of six settings: 15 factual instances per setting, 90 repeated factual instances per target. Setting- and seed-level records remain archived.

\begin{table}[H]
\centering\footnotesize
\caption{Posterior search-target ablation on the original 90 factual instances per target. Evaluation fractions condition on returned actions. New-sample validity percentages are computed from pooled passed/returned counts, not averaged across seeds.}
\label{tab:gp-posterior-tau}
\setlength{\tabcolsep}{3pt}\renewcommand{\arraystretch}{1.08}

\begin{tabular*}{\linewidth}{@{\extracolsep{\fill}}lrrrrrr@{}}\toprule
& \multicolumn{3}{c}{\cref{alg:worst-case}} & \multicolumn{3}{c}{\cref{alg:weighted}} \\
\cmidrule(lr){2-4}\cmidrule(l){5-7}
$\tau$ & Returned & \shortstack[r]{New-sample validity\\passed/returned (\%)} & True/ret. & Returned & \shortstack[r]{New-sample validity\\passed/returned (\%)} & True/ret. \\ \midrule
0.5 & 68/90 & 63/68 (92.6\%) & 67/68 & 74/90 & 66/74 (89.2\%) & 61/74 \\
0.7 & 68/90 & 64/68 (94.1\%) & 67/68 & 75/90 & 73/75 (97.3\%) & 65/75 \\
0.8 & 68/90 & 64/68 (94.1\%) & 67/68 & 75/90 & 73/75 (97.3\%) & 67/75 \\
0.9 & 68/90 & 64/68 (94.1\%) & 67/68 & 75/90 & 73/75 (97.3\%) & 67/75 \\
0.93 & 68/90 & 64/68 (94.1\%) & 67/68 & 75/90 & 73/75 (97.3\%) & 67/75 \\
\bottomrule
\end{tabular*}
\end{table}\FloatBarrier
The returned count for \cref{alg:worst-case} is unchanged. \cref{alg:weighted} improves from 61 true successes at $\tau=0.5$ to 67 at $\tau=0.9$. Equal aggregate counts at 0.8--0.93 do not imply identical actions or establish a preferred target on new data.

\endgroup

\section{Unified Algorithm}

\Cref{alg:worst-case,alg:weighted} are special cases of \cref{alg:unified}, which replaces the soft minimum \eqref{eq:softmin} with a weighted soft minimum.  \Cref{alg:worst-case} is recovered with uniform $\pi$, $\varepsilon=0$, and all graphs used at every step; its threshold $\tau$ corresponds to $\tau+\log|\GG|/\beta$ here, since uniform weights shift $\tilde{V}_\beta$ by a constant.  \Cref{alg:weighted} is recovered with $\beta\to0$, where the weighted soft minimum becomes the weighted mean $\hat{V}_\pi$.

This is an algebraic extension, not a separately benchmarked procedure. The sampled weighted-mean limit matches \cref{alg:weighted} when repeated indices reuse a common per-graph estimate. At finite positive $\beta$, the sampled log-sum-exp is generally biased for the full weighted soft minimum. The hard acceptance condition remains separate from the search aggregator.

\begin{algorithm2e}[H]
\caption{Unified graph-weighted search with verified return}
\label{alg:unified}
\KwIn{Factual $\xF$; classifier $h$; GP-SCMs $\{\mathcal{M}_{\mathcal{G}}\}_{\mathcal{G}\in\GG}$; graph weights $\pi$; permitted failure weight $\varepsilon$; target sets $\mathfrak{I}$; feasible values $\Theta_{\mathcal{I}}$; cost $\costF$; graphs per step $K$ (or ``all''); parameters $\eta,\lambda,\kappa,\beta,\tau,\gamma_{\textsc{lcb}},T,M,M'$.}
\KwOut{An action $a^\star$, or ``no action found''.}
Set $a^\star$ to empty and its cost to $+\infty$\;
\ForEach{$\mathcal{I}\in\mathfrak{I}$ and initialization $\btheta\in\Theta_{\mathcal{I}}$}{
 \For{$t=0,\ldots,T$}{
  Form $a=do(\XI=\btheta)$ and compute $\hat{\Delta}_{M'}(a;\mathcal{G})$ on fresh samples for every $\mathcal{G}\in\GG$\;
  \If{$\hat{R}_\pi(a)\geq1-\varepsilon$}{
   Keep $a$ if its cost is lower than that of $a^\star$\;
  }
  \If{$t<T$}{
   \eIf{$K=$ ``all''}{
    Use every $\mathcal{G}_k\in\GG$ with weight $w_k=\pi_{\mathcal{G}_k}$\;
   }{
    Draw $\mathcal{G}_1,\ldots,\mathcal{G}_K\sim\pi$ independently, each with weight $w_k=1/K$\;
   }
   Compute one $\hat\Delta_M$ and $V_{\mathcal G,\kappa}$ per distinct evaluated graph, reusing the estimates for repeated indices\;
   Compute $\tilde{V}_{\beta,\pi}(a)=-\frac1\beta\log\sum_k w_k\exp\big(-\beta\,V_{\mathcal{G}_k,\kappa}(a)\big)$, or $\sum_k w_k V_{\mathcal{G}_k,\kappa}(a)$ if $\beta=0$\;
   Compute $\mathcal{L}(\btheta)=\costF(a)+\lambda\big[\tau-\tilde{V}_{\beta,\pi}(a)\big]_+$\;
   Update $\btheta\leftarrow\Pi_{\Theta_{\mathcal{I}}}(\btheta-\eta\nabla_{\btheta}\mathcal{L})$\;
  }
 }
}
Return $a^\star$ if found; otherwise return ``no action found''\;
\end{algorithm2e}

\section{Axes of robustness in algorithmic recourse}
\label{app:robustness-axes}

\cref{fig:robustness-axes} contrasts causal graph ambiguity with other sources of uncertainty considered in robust recourse.

\begin{figure}[t]
\centering
\begin{lrbox}{\revisiontable}% Five axes of robustness in algorithmic recourse (adapted from the proposal figure).
% Requires tikz libraries: arrows.meta. Uses \xF and \xCF from the preamble.
\begin{tikzpicture}[x=1cm,y=1cm,>={Stealth[length=2.2mm]},
 every node/.style={font=\normalsize},
 lab/.style={font=\footnotesize\bfseries,align=center},
 v/.style={circle,draw,fill=white,minimum size=2.2mm,inner sep=0pt}]

% --- Decision regions and boundaries (model multiplicity) ---
\node[lab,font=\normalsize\bfseries,text=red!70!black] at (2.6,6.5) {Denied};
\node[lab,font=\normalsize\bfseries,text=green!45!black] at (11.4,6.5) {Approved};
\draw[gray,line width=1pt] (5.9,0.8) .. controls (6.6,3.0) and (7.4,5.0) .. (8.7,6.6)
 node[above,font=\footnotesize,text=gray] {$h$};
\draw[gray,dashed,line width=.8pt] (4.7,0.9) .. controls (5.4,3.0) and (6.1,5.0) .. (7.2,6.6);
\draw[gray,dotted,line width=1pt] (7.1,0.9) .. controls (7.9,2.8) and (8.8,4.6) .. (10.2,6.0);
\node[lab,text=gray!80!black] at (5.0,7.5) {Model multiplicity\\[-1pt]$\{h_m\}_{m=1}^{M}$};

% --- Noisy representation around the factual ---
\coordinate (xf) at (2.8,2.7);
\draw[red!70!black,dashed] (xf) circle[radius=0.8];
\fill[orange] (xf) circle[radius=1.3mm];
\node[font=\footnotesize,anchor=north] at ([yshift=-1.5mm]xf) {$\xF$};
\node[lab,text=red!70!black] at (2.8,1.25) {Noisy representation\\[-1pt]$\xF+\epsilon$};

% --- Action and noisy execution ---
\coordinate (xs) at (10.0,3.5);
\draw[blue!70!black,dashed] (xs) circle[radius=1.0];
\draw[->,line width=1pt] (xf) -- (xs)
 node[pos=0.25,above=2pt,font=\normalsize] {$a$};
\fill[blue!60] (xs) circle[radius=1.3mm];
\node[font=\footnotesize,anchor=north] at ([yshift=-1.5mm]xs) {$\xCF(a)$};
\node[lab,text=blue!70!black,anchor=west] at (11.1,3.5) {Noisy execution\\[-1pt]$\xCF(a+\delta)$};

% --- Causal ambiguity: one edge whose orientation the MEC leaves open ---
\node[v] (g1) at (0.6,4.6) {};
\node[v] (g2) at (1.8,4.6) {};
\node[v] (g3) at (1.2,5.5) {};
\draw[->] (g1) -- (g2);
\draw[<->,red!70!black,dashed] (g1) -- node[above left=-1pt,font=\scriptsize,text=red!70!black] {?} (g3);
\draw[->] (g3) -- (g2);
\node[lab,text=black!80] at (1.2,3.95) {Causal ambiguity};

% --- Temporal drift ---
\draw[->,violet,line width=1pt] (0.4,0) -- (12.4,0);
\foreach \x in {3,6,9} \draw[violet] (\x,-0.12) -- (\x,0.12);
\node[lab,text=violet,anchor=north east] at (12.4,-0.15) {Temporal drift $h_t$};
\end{tikzpicture}\end{lrbox}
\ifdim1.05\wd\revisiontable>\linewidth\resizebox{\linewidth}{!}{\usebox{\revisiontable}}\else\scalebox{1.05}{\usebox{\revisiontable}}\fi
\caption{Five axes of robustness in algorithmic recourse.  An action $a=do(\XI=\btheta)$ moves the factual $\xF$ from the denied to the approved side of the classifier $h$.  (1) \emph{Noisy representation}: the recorded factual may differ from the true one, $\xF+\epsilon$.  (2) \emph{Noisy execution}: the individual may implement a perturbed action, $\xCF(a+\delta)$.  (3) \emph{Model multiplicity}: several classifiers $\{h_m\}$ may fit the data equally well, with conflicting decision boundaries.  (4) \emph{Temporal drift}: the classifier $h_t$ and data distribution change over time.  (5) \emph{Causal ambiguity}, the focus of this paper: observationally equivalent causal graphs disagree on edge orientations, and hence on the downstream consequences of the same action.}
\label{fig:robustness-axes}
\end{figure}
\FloatBarrier

\end{document}